\documentclass{article}

\usepackage{microtype}
\usepackage{graphicx}
\usepackage{subcaption}
\usepackage{booktabs} 

\usepackage{hyperref}

\usepackage[preprint]{icml2026}

\usepackage{amsmath}
\usepackage{amssymb}
\usepackage{mathtools}
\usepackage{multirow}
\usepackage{colortbl}
\usepackage{booktabs} 
\usepackage{pifont}    

\usepackage{rotating}

\definecolor{demphcolor}{gray}{.2}

\definecolor{demphcolorinline}{gray}{.3}

\definecolor{demphcolor1}{gray}{.6}

\newcommand{\methodname}{CaMo-3B}
\newcommand{\datasetname}{CaMo-30K}

\definecolor{LightCyan}{rgb}{0.94, 1.0, 1.0}
\definecolor{ForestGreen}{rgb}{0.13, 0.55, 0.13}
\definecolor{LightBlue}{rgb}{0.678, 0.847, 0.902}
\definecolor{DarkRed}{rgb}{0.7, 0.1, 0.1}

\newcolumntype{C}[1]{>{\centering\arraybackslash}p{#1}}
\newcolumntype{L}[1]{>{\arraybackslash}p{#1}}

\usepackage[capitalize,noabbrev]{cleveref}

\usepackage[textsize=tiny]{todonotes}

\icmltitlerunning{Camera Motion Grounded Evaluation and Training for Vision-Language Models}

\begin{document}

\twocolumn[
  \icmltitle{CaMo: Camera Motion Grounded Evaluation and Training \\for Vision-Language Models}




\begin{icmlauthorlist}
\icmlauthor{Hsiang-Wei Huang}{uw}
\icmlauthor{Junbin Lu}{uw}
\icmlauthor{Kuang-Ming Chen}{uw}
\icmlauthor{Jianxu Shangguan}{uw}
\icmlauthor{Cheng-Yen Yang}{uw}
\icmlauthor{Jenq-Neng Hwang}{uw}
\end{icmlauthorlist}

\icmlaffiliation{uw}{Department of Electrical and Computer Engineering, University of Washington, Seattle, USA}

\icmlcorrespondingauthor{Jenq-Neng Hwang}{hwang@uw.edu}

  \icmlkeywords{Spatial Understanding}

  \vskip 0.3in
]



\printAffiliationsAndNotice{}  

\begin{abstract}

Vision-Language Models~(VLMs) achieve strong performance on spatial question answering benchmarks, yet it remains unclear whether such gains reflect genuine spatial intelligence. We show that existing spatial VLMs lack basic camera motion understanding, a key component of spatial cognition. We propose the Spatial Narrative Score (SNS), an evaluation framework that requires VLMs to generate explicit spatial narratives capturing both scene semantics and camera motion, followed by reasoning with a frozen proxy LLM. Under SNS, state-of-the-art spatial VLMs exhibit significant performance degradation despite high direct question answering accuracy. To address this gap, we introduce CaMo, a camera motion grounded VLM that achieves consistent performance across SNS evaluation and direct spatial question answering accuracy. Our results highlight the importance of explicit spatial narrative externalization for evaluating VLMs with transferable 3D spatial understanding. Our code, data, and model is available at \href{https://github.com/hsiangwei0903/CaMo}{https://github.com/hsiangwei0903/CaMo}.
\end{abstract}
\section{Introduction}
Vision-Language Models~(VLMs) have achieve remarkable success in visual question answering~\citep{li2024llava-nxt,li2024llava-one,tong2024cambrian}, captioning~\citep{lian2025describe,chen2024sharegpt4video}, and grounding~\citep{bai2025qwen2,wang2025learning}. Building on these advances, recent research has increasingly focused on extending VLMs beyond recognition-centric tasks toward spatial intelligence. To this end, prior work has explored diverse directions, including introducing new architectural components~\citep{wu2025spatial,zhao2025spacemind,fan2025vlm3r} or enhancing spatial reasoning through reinforcement learning method such as Group Relative Policy Optimization (GRPO)~\citep{ouyang2025spacer,li2025spatialladder}. Notably, the introduction of GRPO~\citep{shao2024deepseekmath,guo2025deepseek} has emerged as a dominant strategy and achieve substantial gains on spatial question answering accuracy~\citep{yang2025thinking,li2025spatialladder}.

\begin{figure}[!t]
\centering
\includegraphics[width=0.98\linewidth]{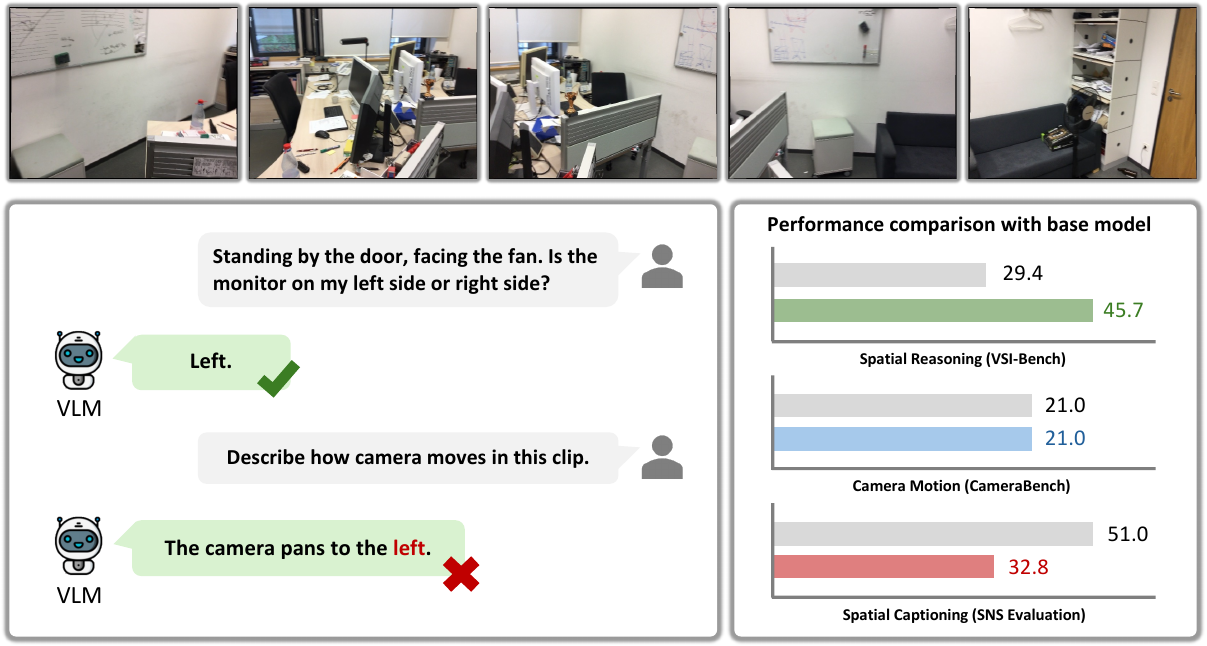}
    \caption{Fine-tuning on spatial QA data improves VLM's~\citep{li2025spatialladder} spatial question answering, but does not translate to better camera motion understanding compared to base model.}
    \label{fig:teaser}
\end{figure}

From a human perspective, spatial cognition fundamentally relies on egomotion signals provided by the vestibular system~\citep{day2005vestibular}, which enable the brain to estimate self-motion and construct spatial representation of the surrounding world. In contrast, VLMs lack access to any biological signals and must rely on visual input to understand egomotion. This limitation makes egomotion inference a critical yet underexplored component of spatial cognition in VLMs. To establish spatial correspondence across views and perform spatial reasoning, a model must first understand how the camera moves through the environment. Despite its importance, existing spatial VLMs' training and evaluation predominantly focus on question answering, and largely overlook understanding the camera motion as a foundation for building spatial intelligence.

Despite the improved accuracy achieved by GRPO-based spatial VLMs, our experiments reveal these models do not achieve corresponding improvements in camera motion understanding, as shown in Figure~\ref{fig:teaser}. These findings raise a fundamental question on whether the improvements reflect genuine spatial understanding, or are they a mirage induced by shortcut learning aligned with benchmark-specific answer distributions? To address this, we believe evaluating VLM spatial intelligence should go beyond answer accuracy and examine whether a model possesses the foundational capability to infer camera motion from visual input, as it is a prerequisite for establishing spatial correspondence and constructing a coherent spatial representation. If a model fails to infer camera motion from video, it cannot reliably reason about spatial structure, regardless of its performance on downstream question answering task. Therefore, assessing camera motion understanding serves as a principled diagnostic for distinguishing genuine spatial reasoning from shortcut-driven behavior.

In this work, we propose a novel evaluation method for assessing spatial understanding in VLMs. Given existing spatial question answering benchmarks with fixed video–question pairs, we prompt a VLM to generate dense spatial narratives that explicitly capture both the semantic content and the camera motion of the video, and feed the generated narrative together with the benchmark question into an off-the-shelf, proxy Large Language Model~(LLM) that performs explicit chain-of-thought reasoning to infer the final answer. This evaluation paradigm, termed Spatial Narrative Score~(SNS), decouples spatial perception from answer prediction and mitigates shortcut learning by requiring models to produce accurate and informative spatial narratives grounded in camera motion and visual semantics. Using SNS, we evaluate the SOTA spatial VLM SpatialLadder~\citep{li2025spatialladder} and its base model Qwen2.5-VL~\citep{bai2025qwen2}, and observe a surprising phenomenon: while SpatialLadder achieved improved accuracy on spatial question answering, it leads to a significant degradation under SNS evaluation, as illustrated in Figure~\ref{fig:teaser}. This discrepancy suggests that improvements from spatial reasoning supervision do not necessarily translate to enhanced camera motion understanding, but may instead arise from shortcut learning aligned with benchmark answer distributions.

Motivated by these findings, we argue that explicitly learning to infer camera motion is a critical step toward genuine spatial intelligence, and therefore construct a new training dataset \datasetname, which integrates video camera motion captioning with spatial question answering data. By performing vanilla supervised fine-tuning without adopting additional modality, architectural components, or GRPO training, we introduce \methodname, a spatial understanding VLM that achieves consistent performance across spatial narrative score evaluation, spatial question answering, and camera motion captioning benchmarks. Our results validates the effectiveness of our training framework in improving VLM's ability in spatial understanding. We summarize our main contributions as follows:

\begin{itemize}
    \item We propose Spatial Narrative Score (SNS), a diagnostic evaluation method that assesses spatial understanding by decoupling spatial perception from answer prediction via narrative generation and proxy LLM reasoning.
    
    \item Using SNS, we reveal that SOTA spatial understanding VLMs can achieve high benchmark accuracy without understanding camera motion, exposing the limitation of existing question answering evaluation in assessing VLM's genuine spatial intelligence.
    
    \item Build upon our findings, we further introduce~\methodname, a spatial understanding VLM with advanced and consistent performance on both direct spatial question answering accuracy and SNS evaluation.
\end{itemize}

\section{Related Work}
\begin{figure*}[!t]
\centering
\includegraphics[width=0.98\linewidth]{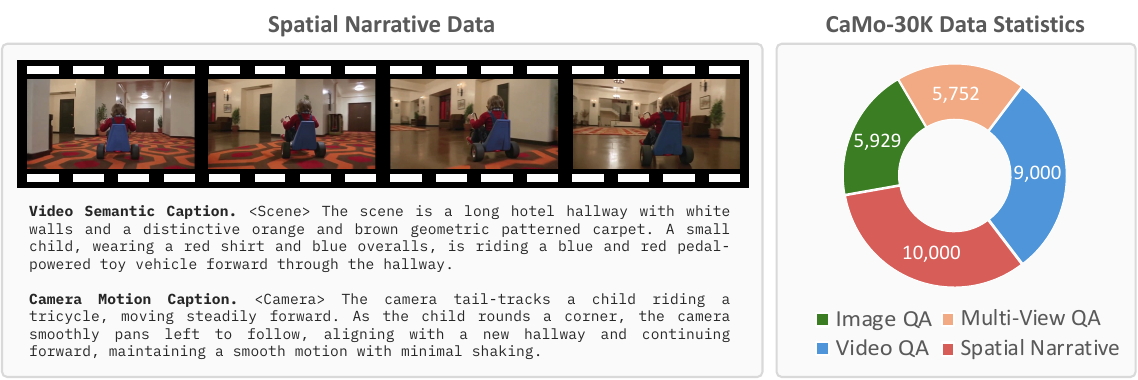}
    \caption{\textbf{Left.} \datasetname~features detailed video semantic and camera motion caption in a structured format. \textbf{Right.} Our data composes a mixture of image, multi-view, and video spatial understanding QA pairs and our spatial narrative data, with a total of 30K samples.}
    \label{fig:data}
\end{figure*}

\subsection{Spatial Understanding VLM}
While modern VLMs achieve advancement on general visual understanding~\citep{li2024llava-one, li2024llava-nxt}, they continue to struggle on challenging spatial understanding~\citep{yang2025thinking, zhang2025flatland, li2025viewspatial}. Although GRPO-based spatial VLMs~\citep{liao2025improved, ouyang2025spacer, li2025spatialladder} report notable gains on spatial understanding benchmarks, existing evaluations mainly focus on question answering accuracy and provide limited insight into whether models acquire a coherent understanding of camera motion—a key signal for building spatial representations from visual input. Our work shows that SOTA spatial VLMs struggle to accurately describe camera motion from video, suggesting that question answering accuracy improvements does not fully reflect generalizable spatial intelligence.

\subsection{Camera Motion Understanding}
Camera motion understanding plays a critical role in building spatial representation. Recent studies incorporate camera motion annotations into controllable video generation models~\citep{yang2024direct, wu2024motionbooth}, highlighting it as an important supervisory signal. Several video captioning VLMs~\citep{wang2024tarsierrecipestrainingevaluating, chaiauroracap} can describe coarse camera movements, but typically fail to capture fine-grained motion patterns such as arcing or tracking behaviors, with only CameraBench~\citep{camerabench} represents an early effort to annotate such fine-grained camera motion data. Distinct from prior work, we leverage camera motion understanding as a foundation for improving spatial understanding of VLMs. Building upon CameraBench, we enrich the data with dense semantic video captions and integrate camera motion with visual content into unified spatial narratives, enabling camera motion grounded supervision for spatial understanding VLM training.
\section{\datasetname~Dataset}
\subsection{Dataset Construction}
Inspired by human spatial cognition, which relies on the integration of camera motion and visual perception to construct a consistent 3D understanding of the environment, we identify these two factors as fundamental to enabling spatial intelligence in Vision-Language Models. As VLMs lack access to biological egomotion signals, camera motion inferred from visual observations serves as the primary clue for building spatial representation. To this reason, our objective is to construct a dataset that explicitly supervises VLMs to infer both camera motion and visual semantic content from video, thereby strengthening their ability to reason about spatial layout. 

To this end, we collect video–camera motion caption pairs from CameraBench~\cite{camerabench}, which contains 3K videos with human-annotated fine-grained camera motion descriptions. Since CameraBench does not include semantic scene captions, we employ Gemini-2.5-Flash~\cite{comanici2025gemini} to generate dense semantic captions for each video. The semantic captions and camera motion descriptions are integrated into a unified spatial narrative format using special scene and camera tag, as illustrated in Figure~\ref{fig:data}. To improve robustness and reduce prompt overfitting, we further augment the data using multiple input prompt templates. This process yields 10K video-based spatial narrative training samples.

To further enhance spatial understanding beyond narrative supervision, we additionally incorporate spatial question answering data into our training data, including 11K single- and multi-view QA pairs from SpatialLadder~\cite{li2025spatialladder} and 9K video QA pairs from SpaceR~\cite{ouyang2025spacer}. All the 3D scenes in the testing benchmarks are filtered from the training data. The resulting dataset comprises 30K samples of spatial narrative generation and spatial question answering data, which we term \textbf{\datasetname} to reflect its objective of enhancing \textbf{Ca}mera \textbf{Mo}tion understanding in VLMs. An overview of the dataset is illustrated in Figure~\ref{fig:data}.

\begin{figure*}[!t]
\centering
\includegraphics[width=0.98\linewidth]{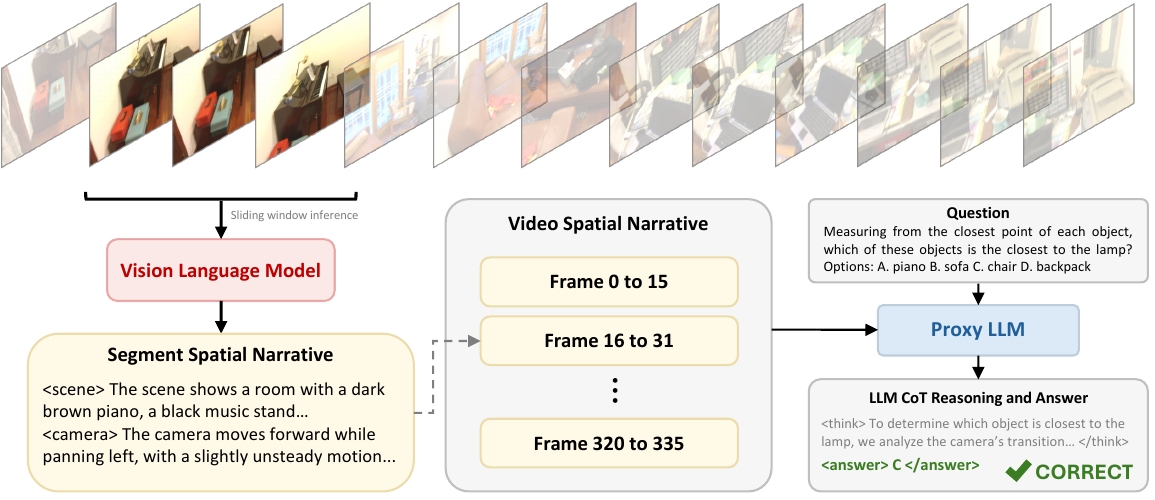}
    \caption{\textbf{Pipeline of Spatial Narrative Score Evaluation.} The input video segment will be sent to the VLM to generate dense video spatial narrative, which is sent to an proxy LLM to generate prediction for accuracy evaluation.}
    \label{fig:sns}
\end{figure*}

\subsection{Quality Control}

To ensure the quality and reliability of \datasetname, rigorous quality control are adopted from both the original data sources and our own verification process. This includes data filtering mechanisms for spatial understanding dataset to balance answer distribution, object diversity, and answer visibility. Human-in-the-loop validation is also adopted for camera motion annotations in CameraBench. For our semantic captioning pipeline, we conduct an internal quality verification step on a randomly selected 500 samples from the data by examine the semantic fidelity and motion consistency of video captions. The internal quality check yields a 99.8\% pass rate on these two criterion, which shows the high-quality of our collected captions for VLM training.


\subsection{\methodname~Training}
We follow recent spatial understanding VLMs~\citep{ouyang2025spacer,li2025spatialladder} and adopt the same base model Qwen2.5-VL-3B~\citep{bai2025qwen2}. We perform supervised fine-tuning on \datasetname~and introduce a strong spatial understanding VLM \methodname. Importantly, we retain the original Qwen2.5-VL architecture and do not introduce additional GRPO training~ techniques~\citep{li2025spatialladder,yuan2025scene}, modules~\citep{fan2025vlm3r,zhao2025spacemind}, or auxiliary modalities like prior works~\citep{huang20253d, yuan2025scene}. This allows us to isolate the effect of training data and supervision signals, ensuring that performance improvements can be directly attributed to our dataset rather than external training or architectural advantages. The full training recipes can be found in Table~\ref{table:training_config} in our appendix.

\section{Spatial Narrative Score Evaluation}
\subsection{Shortcut Learning in Spatial Understanding VLMs}
Many recent works demonstrate spatial QA-finetuned VLM can achieve substantial accuracy improvements in spatial question answering benchmark such as VSI-Bench. However, the evidence of these VLMs achieving genuine spatial intelligence beyond accuracy improvements on these benchmark is somewhat lacking. Instead, recent study~\citep{brown2025benchmark} discovers the potential of VLMs exploiting spatial understanding dataset bias through shortcut learning, which raise doubts on whether these VLMs really form spatial intelligence despite their impressive QA accuracy. To this reason, we found it necessary to decouple the objective in training and evaluation stage. In other words, breaking the objective gradient in training and evaluation.

\subsection{Spatial Narrative Score Evaluation}
We introduce the Spatial Narrative Score (SNS), a novel evaluation metrics designed to assess the spatial understanding capability of VLMs beyond direct answer prediction. Spatial Narrative Score focuses on evaluating whether the VLM can explicitly articulate the spatial evidence required to solve a spatial reasoning task, rather than merely producing the correct answer. We illustrate SNS in Figure~\ref{fig:sns}.

Given an input video, we uniformly partition it into fixed-length, non-overlap temporal segments. For each segment, the VLM is prompted to generate a dense spatial narrative, consisting of semantic scene descriptions and explicit camera motion descriptions. The segment-level narratives are then concatenated into a structured video-level spatial narrative. This spatial narrative is subsequently provided, together with the original benchmark question, to an off-the-shelf LLM with strong reasoning capability. Acting as a proxy reasoner, the LLM performs multi-step reasoning over the narrative and predicts the final answer, as illustrated in Figure~\ref{fig:sns}. The VLM itself does not have access to the question and therefore is never asked to directly output the answer during SNS evaluation. By shifting the evaluation target from answer prediction to evidence generation, SNS substantially reduces the possibility that a VLM can exploit benchmark-specific shortcuts. To perform well under SNS, a model must generate narratives that accurately encode both what is observed in the scene and how the camera moves, which together form the basis of spatial cognition.

Under this evaluation framework, a discrepancy between direct question answering and SNS performance provides diagnostic insight. A model that performs well in direct question answering but fails to generate coherent and faithful spatial narratives suggests reliance on shortcut learning rather than explicit spatial reasoning. We integrate SNS with the VSI-Bench for evaluation. Since spatial narratives alone are insufficient to resolve numerical estimation questions (e.g., object size or metric distances), we focus on multiple-choice question~(e.g., relative direction, route planing).
\begin{table*}[t]
\centering
\small
\caption{\textbf{Comparison of SNS Evaluation results on VSI-Bench.} SNS accuracies are reported as percentages (\%), and improvements are relative to the Qwen2.5-VL-3B. \textbf{Bold} indicates best performance, while \underline{underlined} represents second-best.}
\label{tab:vsi-comparison-inline}
\resizebox{\textwidth}{!}{
\begin{tabular}{lcccccc}
\toprule
\textbf{Model} & \textbf{Training} & \textbf{Rel. Dir.} & \textbf{Rel. Dist.} & \textbf{Appr. Order} & \textbf{Route Plan.} & \textbf{Overall} \\
\midrule
Qwen2.5-VL-3B~\citep{bai2025qwen2} & SFT & 38.0 & 38.0 & 57.1 & 44.9 & 44.4 \\
SpatialLadder-3B~\citep{li2025spatialladder} & GRPO & 30.0 (\textcolor{DarkRed}{-8.0}) & 32.0 (\textcolor{DarkRed}{-6.0}) & 38.8 (\textcolor{DarkRed}{-18.3}) & 30.6 (\textcolor{DarkRed}{-14.3}) & 32.8 (\textcolor{DarkRed}{-11.6}) \\
SpaceR-3B~\citep{ouyang2025spacer} & GRPO & \underline{36.0} (\textcolor{DarkRed}{-2.0}) & \underline{44.0} (\textcolor{ForestGreen}{+6.0}) & \textbf{75.5} (\textcolor{ForestGreen}{+18.4}) & 38.8 (\textcolor{DarkRed}{-6.1}) & \underline{48.5} (\textcolor{ForestGreen}{+4.1}) \\
\midrule
\rowcolor{gray!10}
\textbf{\methodname} & SFT & \textbf{40.0} (\textcolor{ForestGreen}{+2.0}) & \textbf{46.0} (\textcolor{ForestGreen}{+8.0}) & \underline{69.4} (\textcolor{ForestGreen}{+12.3}) & \textbf{49.0} (\textcolor{ForestGreen}{+4.1}) & \textbf{51.0} (\textcolor{ForestGreen}{+6.6}) \\
\bottomrule
\end{tabular}
\label{table:sns}
}
\end{table*}
\section{Experiment}
\subsection{Implementation Details}
We adopt Qwen2.5-VL-3B as base VLM and trained on \datasetname~for one epoch with supervised fine-tuning, the full training recipe is listed in Table~\ref{table:training_config} in appendix. For Spatial Narrative Score Evaluation, we adopt Gemini-2.5-flash as the proxy LLM. Each video segment has 16 frames. More evaluation details and used prompt can be found in Section~\ref{sec:evaluation_setting} and Section~\ref{sec:evaluation_prompt} in appendix.

\subsection{Benchmarks}
We evaluate on multiple spatial understanding benchmarks, including in-domain benchmarks VSI-Bench~\cite{yang2025thinking} and SpatialBench~\cite{li2025spatialladder} with similar question types seen in our training data. We also evaluated on out-of-domain benchmarks with unseen question types~(\cite{tong2024cambrian,zhang2025flatland,li2025viewspatial}) to show generalizability. Lastly, we evaluate on the camera motion captioning task using CameraBench~\cite{camerabench}. More benchmark details can be found in appendix, Section~\ref{sec:benchmark}.

\subsection{Spatial Narrative Score Evaluation}
We evaluate recent spatial understanding VLMs SpaceR-3B and SpatialLadder-3B on the multi-choice questions~(MCQ) from the 198 questions of official VSI-Bench tiny set in Table~\ref{table:sns}. Spatial Narrative Score evaluates a model’s ability to generate accurate spatial narratives that jointly capture scene semantics and camera motion, which serve as intermediate representations for spatial reasoning.
Among all models, our \methodname~achieves the highest overall SNS performance.
Below we highlight several key observations.

\paragraph{Strong Performance of the Base Model.}
We first observe that Qwen-2.5-VL already achieves a competitive SNS results, indicating its strong ability to describe visual semantics and coarse camera motion. However, fine-tuning Qwen-2.5-VL on spatial question answering data does not lead to a corresponding improvement under SNS evaluation.


\begin{figure}
\centering
\includegraphics[width=\linewidth]{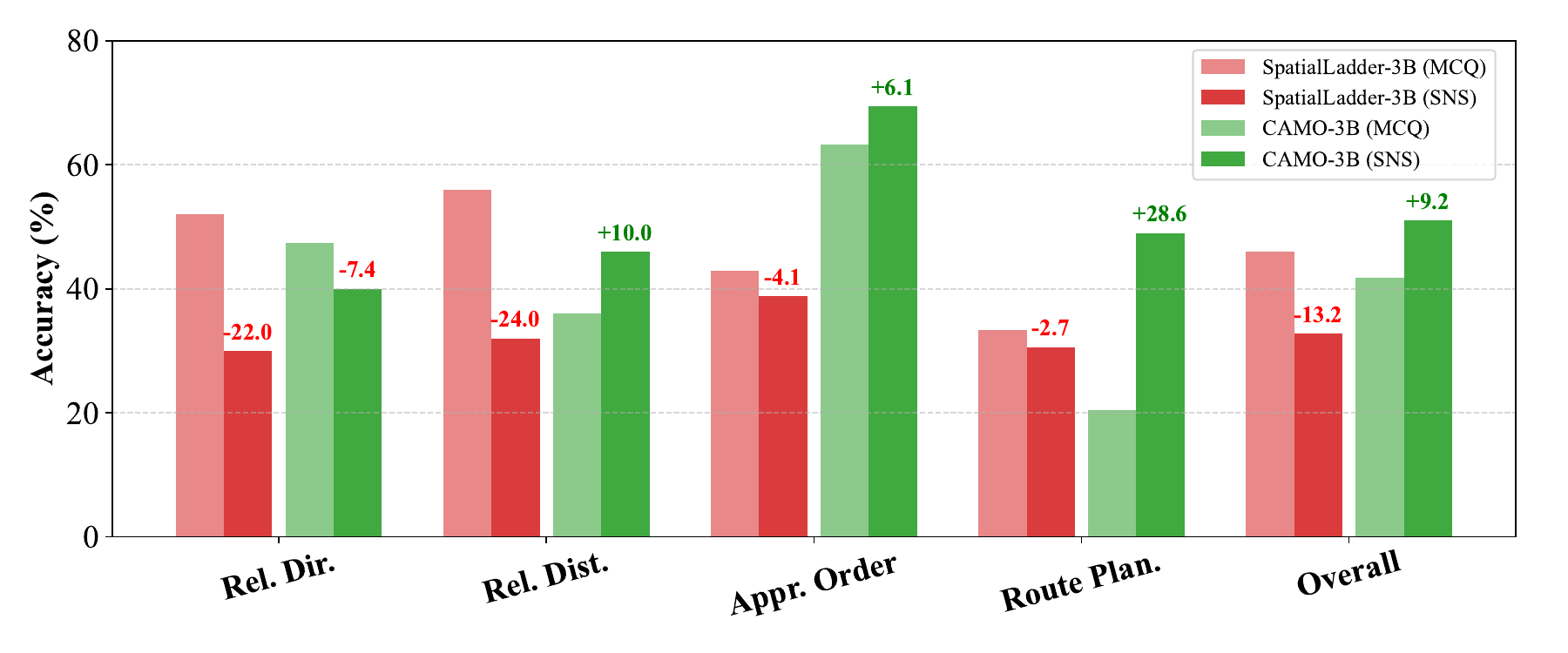}
\caption{\textbf{Gap Between Direct MCQ Accuracy and SNS.} SpatialLadder exhibits a substantial performance drop across all question types under SNS evaluation, whereas CaMo maintains more consistent or even improved performance in all question types.}
\label{fig:vsi-tiny}
\end{figure}

\begin{table*}
\centering
\small
\caption{\textbf{Evaluation Results on In-domain Benchmarks.} NQ and MCQ denotes numerical question and multiple-choice question, respectively. \textbf{Bold} numbers indicate the best performance, while \underline{underlined} numbers represent the second-best performance.}
\resizebox{\textwidth}{!}{  %
\begin{tabular}
{lC{0.8cm}C{0.8cm}C{0.8cm}C{0.8cm}C{0.8cm}C{0.8cm}C{0.8cm}C{0.8cm}C{0.8cm}C{1.2cm}}
\toprule
\multirow{2}{*}{\textbf{Model}} & \multicolumn{3}{c}{\textbf{VSI-Bench}} & \multicolumn{3}{c}{\textbf{SPBench-SI}} & \multicolumn{3}{c}{\textbf{SPBench-MV}} & \multirow{2}{*}{\textbf{Overall}} \\
\cmidrule(lr){2-4} \cmidrule(lr){5-7} \cmidrule(lr){8-10}
& NQ & MCQ & Avg. & NQ & MCQ & Avg. & NQ & MCQ & Avg.& \\
\midrule
\rowcolor{gray!10} 
\multicolumn{11}{l}{\textit{\textbf{Proprietary Models}}} \\
GPT-4o~\citep{hurst2024gpt} & 33.4 & 34.6 & 34.0 & 24.5 & 60.3 & 42.4 & 40.7 & 59.4 & 48.2 & 41.5 \\
Gemini-2.0-Flash~\citep{team2024gemini} & 46.4 & \textbf{44.3} & 45.4 & 49.0 & 60.4 & 54.7 & 51.9 & 50.7 & 56.5 & 52.2 \\
\midrule
\rowcolor{gray!10} 
\multicolumn{11}{l}{\textit{\textbf{Open-Source Models}}} \\
InternVL-2.5-4B~\citep{chen2024internvl} & 30.6 & 34.1 & 32.6 & 31.8 & 53.3 & 42.5 & 37.7 & 51.4 & 43.2 & 42.8 \\
InternVL-2.5-8B~\citep{chen2024internvl} & 40.4 & 40.0 & 40.2 & 28.3 & 56.3 & 42.3 & 37.3 & 47.5 & 41.4 & 41.4 \\
Kimi-VL-A3B~\citep{team2025kimi} & 31.8 & 25.5 & 28.7 & 25.7 & 44.9 & 35.3 & 23.3 & 57.6 & 37.0 & 36.0 \\
LLaVA-OneVision-7B~\citep{li2024llava-one} & 34.5 & 31.2 & 33.1 & 25.4 & 41.0 & 33.2 & 20.6 & 49.6 & 32.2 & 32.2 \\
\midrule
\rowcolor{gray!10} 
\multicolumn{11}{l}{\textit{\textbf{Qwen2.5-VL-7B Based Spatial Models}}} \\
Qwen2.5-VL-7B~\citep{bai2025qwen2} & 37.1 & 34.6 & 35.8 & 36.3 & 60.5 & 48.4 & 28.9 & 49.8 & 37.3 & 43.9 \\
SpaceR-7B~\citep{ouyang2025spacer} & 47.8 & 41.2 & 44.5 & 35.7 & 61.5 & 48.6 & 63.2 & 53.7 & 59.4 & 50.8 \\
VILASR-7B~\citep{wu2025reinforcing} & 47.4 & \underline{43.4} & 45.4 & 36.6 & 63.7 & 50.2 & 56.2 & \underline{59.6} & 57.6 & 51.1 \\
Video-R1~\citep{feng2025video} & 33.8 & 32.9 & 33.4 & 27.7 & 62.0 & 44.8 & 32.5 & 53.0 & 40.7 & 39.6 \\
\midrule
\rowcolor{gray!10} 
\multicolumn{11}{l}{\textit{\textbf{Qwen2.5-VL-3B Based Spatial Models}}} \\
Qwen2.5-VL-3B~\citep{bai2025qwen2} & 26.0 & 33.0 & 29.4 & 24.3 & 56.2 & 40.3 & 25.6 & 53.2 & 36.6 & 38.8 \\
Spatial-MLLM-4B~\citep{wu2025spatial} & \textbf{51.5} & 43.1 & \textbf{47.3} & 38.1 & 49.3 & 43.7 & \underline{63.7} & 58.9 & 53.1 & 48.0 \\
SpatialLadder-3B~\citep{li2025spatialladder} & \underline{50.8} & 40.5 & \underline{45.7} & \textbf{58.6} & \textbf{81.8} & \textbf{70.2} & \textbf{68.2} & \textbf{75.0} & \textbf{70.9} & \textbf{62.3} \\
\rowcolor{gray!10}
\textbf{\methodname} & 46.5 & 41.6 & 44.0 & \underline{51.5} & \underline{77.5} & \underline{64.4} & 61.2 & \underline{72.8} & \underline{65.8} & \underline{58.1} \\
\textcolor{ForestGreen}{\textit{\textbf{Improvement}}} & \textcolor{ForestGreen}{+20.5} & \textcolor{ForestGreen}{+8.6} & \textcolor{ForestGreen}{+14.6} & \textcolor{ForestGreen}{+27.2} & \textcolor{ForestGreen}{+21.3} & \textcolor{ForestGreen}{+24.1} & \textcolor{ForestGreen}{+35.6} & \textcolor{ForestGreen}{+19.6} & \textcolor{ForestGreen}{+29.2} & \textcolor{ForestGreen}{+19.3}\\
\bottomrule
\end{tabular}
}
\label{tab:in-domain}
\end{table*}
\begin{table*}[t]
\centering
\small
\vspace{15pt}
\caption{\textbf{Evaluation Results on Out-of-domain Benchmarks.} For ViewSpatial-Bench, CP and PP represent Camera-Perspective and Person-Perspective. \textbf{Bold} numbers indicate the best performance, \underline{underlined} numbers represent the second-best performance.}
\resizebox{\textwidth}{!}{
\begin{tabular}{lC{0.7cm}C{0.7cm}C{0.7cm}C{0.7cm}C{1.2cm}C{0.7cm}C{0.7cm}C{0.7cm}C{0.7cm}C{0.7cm}C{0.9cm}}
\toprule
\multirow{2}{*}{\textbf{Model}} & \multicolumn{3}{c}{\textbf{\mbox{CV-Bench}}} & \multicolumn{4}{c}{\textbf{SPAR-Bench}} & \multicolumn{3}{c}{\textbf{\mbox{ViewSpatial-Bench}}} & \multirow{2}{*}{\textbf{Overall}} \\
\cmidrule(lr){2-4} \cmidrule(lr){5-8}\cmidrule(lr){9-11}
& 2D & 3D & Avg. & Low & Medium & High & Avg. & CP & PP & Avg. & \\
\midrule
GPT-4o & 69.4 & \underline{81.3} & 75.4 & \textbf{29.3} & 24.9 & \textbf{45.1} & \textbf{36.4} & 33.7 & 31.5 & 32.6 & \underline{48.1} \\
InternVL-2.5-4B & \underline{73.5} & 75.1 & 74.4 & 25.7 & \underline{29.8} & 35.2 & 30.6 & 40.8 & 35.1 & \underline{37.9} & 47.6 \\
Kimi-VL-A3B & 41.9 & 54.7 & 48.3 & \underline{26.4} & 18.8 & 36.2 & 29.4 & 25.1 & \textbf{41.5} & 33.6 & 37.1 \\
LLaVA-OneVision-7B & 50.6 & 63.5 & 58.3 & 21.8 & 26.1 & 40.1 & \underline{31.2} & 28.5 & 26.5 & 27.5 & 39.0 \\
Qwen2.5-VL-7B & \textbf{75.0} & \textbf{83.1} & \textbf{79.0} & 17.5 & 29.5 & 41.8 & 30.2 & \textbf{41.8} & 34.3 & \underline{37.9} & \textbf{49.0} \\
\midrule
\rowcolor{gray!10}
Qwen2.5-VL-3B & 69.1 & 72.2 & 70.6 & 15.3 & 26.4 & 32.2 & 24.6 & 39.5 & 32.0 & 35.6 & 43.6 \\
\rowcolor{gray!10}
\methodname & \underline{73.5} & 79.2 & \underline{76.3} & 13.0 & \textbf{31.3} & \underline{42.6} & 27.6 & \underline{41.5} & \underline{40.4} & \textbf{41.0} & 47.4 \\
\rowcolor{gray!10}
\textcolor{ForestGreen}{\textit{\textbf{Improvement}}} & \textcolor{ForestGreen}{+4.4} & \textcolor{ForestGreen}{+7.0} & \textcolor{ForestGreen}{+5.7} & \textcolor{DarkRed}{-2.3} & \textcolor{ForestGreen}{+4.9} & \textcolor{ForestGreen}{+10.4} & \textcolor{ForestGreen}{+3.0} & \textcolor{ForestGreen}{+2.0} & \textcolor{ForestGreen}{+8.4} & \textcolor{ForestGreen}{+5.4} & \textcolor{ForestGreen}{+3.8} \\
\bottomrule
\end{tabular}
}
\label{tab:out-of-domain}
\end{table*}

\paragraph{Severe Degradation in SpatialLadder.}
A striking observation is that SpatialLadder suffers an overall SNS drop of 11.6\% compared to its base model.
This degradation suggests that, although SpatialLadder achieves higher direct-answer accuracy on spatial QA benchmarks, it loses the ability to generate accurate and consistent spatial narratives, particularly regarding camera motion.
As illustrated in Figure~\ref{fig:vsi-tiny}, this performance gap persists across all question categories when comparing SNS with direct MCQ accuracy. This discrepancy reveals a paradoxical behavior: the model can correctly answer complex spatial questions while failing to describe the underlying camera motion that is essential for constructing a coherent global spatial representation. Such a mismatch strongly suggests that the model’s improved benchmark performance is not grounded in robust spatial understanding, but instead relies on shortcut strategies that bypass explicit camera motion understanding.

\paragraph{Data Diversity Mitigates Shortcut Learning.}
Interestingly, SpaceR does not exhibit a similarly severe SNS degradation, despite also adopting GRPO-based training. A key difference lies in its training data composition: SpaceR is trained on 151K samples, among which approximately 60K are general image and video question answering data. We hypothesize that this broader data distribution regularizes the learning dynamics, preventing the model from overfitting to narrow spatial QA patterns and thus mitigating shortcut learning behavior. This observation suggests that shortcut learning is not an inevitable consequence of GRPO itself, but is strongly influenced by data diversity.

\paragraph{SOTA Results of \methodname.} Despite only trained on 30K data~(20\% of SpaceR), \methodname~leverages the camera and semantic video captioning data supervision to achieve strong spatial understanding and camera motion learning, being the only VLMs that establish accuracy improvements over Qwen2.5-VL-3B across all question types in Table~\ref{table:sns}. This suggests our data effectively enhances the model's reliability in generating intermediate representations, rather than merely improving answer-level accuracy.

\begin{table*}[t]
\centering
\caption{Performance on CameraBench. We follow CameraBench evaluation and adopt standard language metrics.}
\resizebox{0.95\linewidth}{!}{
\label{tab:camerabench}
\begin{tabular}{lcccc}
\toprule[1pt]
\multirow{2}{*}{\textbf{Model}} & \multicolumn{4}{c}{\textbf{Caption Generation}} \\
\cmidrule(l){2-5}
& SPICE & ROUGE-L & BLEU-2 & METEOR \\
\midrule
mPLUG-Owl3-7B~\citep{ye2024mplugowl3longimagesequenceunderstanding} & 0.22 & 0.20 & 0.08 & 0.19\\
LLaVA-Video-7B~\citep{llavavideo} & 0.23 & \underline{0.23} & \underline{0.12} & 0.19\\
LLaVA-OneVision-7B~\citep{li2024llava-one} & 0.22 & 0.21 & 0.10 & 0.20\\
InternVideo2-Chat-8B~\citep{wang2024internvideo2} & 0.22 & 0.21 & 0.11 & 0.19\\
Tarsier-Recap-7B~\citep{wang2024tarsierrecipestrainingevaluating} & 0.23 & 0.22 & 0.11 & 0.20\\
InternLMXComposer2.5-7B~\citep{dong2024internlm} & 0.21 & 0.19 & 0.08 & 0.19\\
InternVL2.5-8B~\citep{chen2024internvl} & 0.20 & 0.10 & 0.04 & 0.21\\
InternVL2.5-26B~\citep{chen2024internvl} & 0.23 & 0.20 & 0.09 & 0.23\\
InternVL3-8B~\citep{zhu2025internvl3} & 0.20 & 0.15 & 0.05 & 0.17\\
InternVL3-78B~\citep{zhu2025internvl3} & 0.18 & 0.16 & 0.06 & 0.18\\
Qwen2.5-VL-7B~\citep{bai2025qwen2} & 0.18 & 0.12 & 0.05 & 0.28\\
Qwen2.5-VL-32B~\citep{bai2025qwen2} & 0.24 & 0.17 & 0.08 & 0.29\\
Qwen2.5-VL-72B~\citep{bai2025qwen2} & \underline{0.25} & 0.19 & 0.10 & \underline{0.30}\\
GPT-4o~\citep{hurst2024gpt} & 0.20 & 0.16 & 0.06 & 0.25\\
Gemini-2.5-Flash~\citep{comanici2025gemini} & 0.24 & 0.21 & 0.10 & 0.22 \\
Gemini-2.5-Pro~\citep{comanici2025gemini} & 0.20 & 0.15 & 0.06 & 0.27 \\
\midrule

\rowcolor{gray!10}
Qwen2.5-VL-3B~\citep{bai2025qwen2} & 0.21 & 0.18 & 0.08 & 0.24 \\

\rowcolor{gray!10}
SpaceR-3B~\citep{ouyang2025spacer} & 0.23 & 0.20 & 0.09 & 0.25 \\

\rowcolor{gray!10}
SpatialLadder-3B~\citep{li2025spatialladder} & 0.21 & 0.17 & 0.07 & 0.25 \\

\rowcolor{gray!10}
\textbf{\methodname} & \textbf{0.41} & \textbf{0.38} & \textbf{0.24} & \textbf{0.37}\\
\rowcolor{gray!10}
\textcolor{ForestGreen}{\textit{\textbf{Improvement}}} & \textcolor{ForestGreen}{+0.20} & \textcolor{ForestGreen}{+0.20} & \textcolor{ForestGreen}{+0.16} & \textcolor{ForestGreen}{+0.13} \\
\bottomrule[1pt]

\end{tabular}
}
\end{table*}

\subsection{In-Domain Benchmarks Evaluation}
Table~\ref{tab:in-domain} presents a comprehensive evaluation on several in-domain spatial reasoning benchmarks using their official evaluation settings, including VSI-Bench~\citep{yang2025thinking} and SPBench~\citep{li2025spatialladder}. Overall, \methodname~demonstrates competitive performance against both open-source and proprietary SOTA VLMs. On VSI-Bench, \methodname~achieves competitive accuracy on multi-choice questions~(MCQ), indicating it is able to perform effective spatial reasoning under standard evaluation. For numerical questions~(NQ), our performance is slightly lower than some specialized spatial understanding VLMs. We hypothesize that this gap may be partially attributed to dataset bias in numerical prediction tasks, which recent study have shown to be particularly susceptible to shortcut learning in VSI-Bench~\citep{brown2025benchmark}. Despite this, \methodname~achieves strong overall performance while simultaneously maintaining a clear advantage in generating high-quality spatial narratives, as evidenced by our SNS evaluation. This highlights our VLM's ability in balancing benchmark accuracy and faithful spatial understanding, rather than relying on shortcut-driven numerical cues.

\subsection{Out-of-Domain Benchmarks Evaluation}
We present evaluation results on out-of-domain benchmarks~\citep{tong2024cambrian,zhang2025flatland,li2025viewspatial} in Table~\ref{tab:out-of-domain}, which focuses on question types that are unseen in our training data. Our \methodname~demonstrates robustness and spatial intelligence on out-of-domain questions, with notable performance improvements over base model Qwen-2.5-VL-3B across most question categories. Notably, while our training emphasizes high-level spatial reasoning and narrative grounding, we observe a minor performance drop (2.3\%) on low-level geometric perception tasks in SPAR-Bench. We expect incorporating additional low-level geometric supervision could further improve performance without compromising the strengths of our approach.

\subsection{Camera Motion Captioning Evaluation}
To directly assess VLM's ability in generating accurate camera motion caption, we evaluate the camera motion captioning task on CameraBench~\citep{camerabench} in Table~\ref{tab:camerabench}. \methodname's strong results demonstrating its ability in generating high-quality and accurate camera motion caption. Specifically, \methodname~outperforms other open-source and proprietary VLMs across all evaluation metrics. Furthermore, spatial understanding specialized VLMs like SpaceR and SpatialLadder do not achieve notable performance improvements~(or even degradation) after training on spatial understanding data, suggesting their training strategy does not enable them with improved camera motion understanding, which is an essential component of spatial intelligence.




\begin{table}[t]
\centering
\caption{Ablation on Scene and Camera Captions.}
\resizebox{0.98\linewidth}{!}{
\label{tab:ablation_study}
\begin{tabular}{@{}cccccc@{}}
\toprule
& \textbf{Scene Caption} & \textbf{Camera Caption} & \textbf{Rel Dir.} & \textbf{Route Plan.} & \\ 
\midrule
& \checkmark &  & 36.0 & 42.8 \\
& \checkmark & \checkmark & 40.0~(+4.0) & 49.0~(+6.2) & \\
\bottomrule
\end{tabular}
}
\end{table}

\begin{table}[t]
\centering
\caption{Ablation on segment length in SNS evaluation.}
\label{tab:segment_length_ablation}
\resizebox{0.8\linewidth}{!}{
\begin{tabular}{ccc}
\toprule
\textbf{Frames per Seg.} & \textbf{Num. of Segments} & \textbf{SNS}\\
\midrule
16 & 12.4 & \textbf{51.0} \\
24 & 9.7  & 49.0 \\
32 & 7.4  & 47.0 \\
\bottomrule
\end{tabular}
}
\end{table}


\begin{figure*}[!t]
\centering
\includegraphics[width=0.98\linewidth]{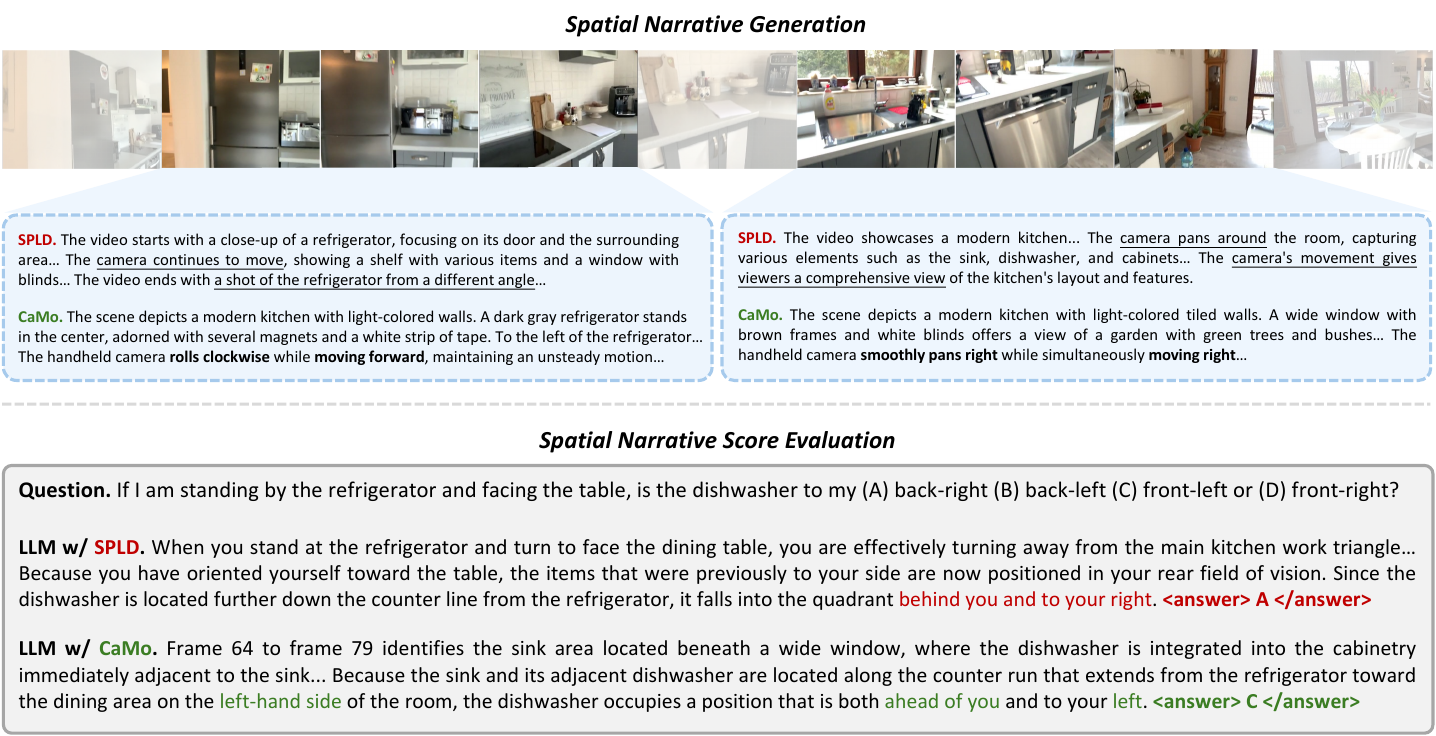}
    \caption{\textbf{Qualitative Results from VSI-Bench.} Compared with the coarse camera motion~(highlighted \underline{underline}) from SpatialLadder, CaMo generates more fine-grained camera motion~(highlighted \textbf{bold}), which contains richer spatial information for the SNS evaluation.}
    \label{fig:qual}
\end{figure*}

\subsection{Ablation Study}

\paragraph{Importance of Camera Caption.} In Table~\ref{tab:ablation_study}, we study the importance of the semantic scene and camera motion caption. The LLM tend to perform better on camera-motion related questions including relative direction~(+4.0) and route planning~(+6.2) when provided with the camera motion. When the camera motion is not available, we observe the LLM tends to rely on the scene semantic caption clue, which also captures dense spatial information that enables the LLM to maintain its performance on question answering.

\paragraph{Segment Length.}
Table~\ref{tab:segment_length_ablation} examines the effect of segment length in SNS evaluation. As segment length increases, each segment aggregates more complex visual content and camera motion, making it harder for the proxy LLM to fully leverage the resulting spatial narrative for question answering. Consequently, SNS performance gradually decreases, indicating that shorter segments better preserve dense and interpretable spatial cues for downstream reasoning. Therefore, we adopt 16 frames per segment as our default evaluation setting for SNS.

\paragraph{Human Spatial Narrative Evaluation.}
To validate that SNS performance is primarily determined by the quality of spatial evidence rather than limitations of the proxy reasoner, we conduct a human spatial narrative evaluation. We randomly sample a subset~(14\%) of challenging questions from the question sets which both CaMo-3B and SpatialLadder fail under SNS evaluation, and replace model-generated narratives with high-quality spatial narratives written by a human annotator. Using the same proxy LLM, these human-authored narratives achieve 70\% accuracy on these challenging questions where spatial understanding VLMs failed. This result indicates that the evaluated questions are solvable given faithful spatial narratives and that SNS failures predominantly stem from insufficient or inaccurate spatial evidence generation, supporting SNS as a meaningful and well-posed evaluation metric for spatial understanding.

\subsection{Qualitative Results} In Figure~\ref{fig:qual}, we present a detailed qualitative results of the spatial narrative generated by CaMo and SpatialLadder~(SPLD). CaMo is able to generate spatial narrative with more fine-grained camera description~(e.g.~rolls clockwise, pans right). While SPLD leans to generate general description~(e.g. continues to move) that does not contain enough spatial information for accurate LLM proxy question answering. CaMo's richer spatial narrative provides the proxy LLM better clue for understanding the spatial layout and generates consistent prediction with the model. Additional analysis including case study~(Section~\ref{sec:casestudy}), word cloud analysis from different VLMs~(Section~\ref{sec:wordcloud}) and failure case analysis~(Section~\ref{sec:failurecase}) can be found in our appendix.
 
\section{Conclusion}
We revisit spatial understanding in Vision–Language Models and show that strong benchmark performance does not necessarily reflect genuine spatial intelligence. To address this gap, we introduce \datasetname, which supervises egocentric camera motion and dense semantic narratives, and propose Spatial Narrative Score~(SNS), an evaluation protocol that decouples perception from reasoning to reduce shortcut exploitation. Our model \methodname~demonstrate that grounding VLMs in temporally coherent motion and semantic evidence leads to more reliable spatial understanding and reveals limitations of existing approaches, offering a clearer direction for evaluating spatial VLMs.
\section*{Impact Statement}
This paper presents work whose goal is to advance the field of Machine
Learning. There are many potential societal consequences of our work, none
which we feel must be specifically highlighted here.

\bibliography{main}
\bibliographystyle{icml2026}

\newpage
\appendix

\onecolumn

\section{Additional Implementation Details}
\label{sec:implementation}
\subsection{\datasetname~Collection Details}
\label{sec:collection-details}
CaMo-30K is constructed to explicitly supervise both semantic scene understanding and camera egomotion inference. The dataset consists of approximately 30K training samples spanning spatial narrative generation and spatial question answering. For spatial narrative data, we collect around 3K videos from CameraBench and expand them into 10K video-level spatial narrative samples using multiple prompt templates. Each spatial narrative contains two components: a dense semantic scene description and a fine-grained camera motion description. Camera motion captions are taken directly from the original CameraBench annotations, while semantic scene captions are generated using Gemini-2.5-flash. 

To complement narrative supervision, we incorporate spatial question answering data from existing benchmarks, including approximately 11K samples from SpatialLadder and 9K samples from SpaceR, covering image-based, multi-view, and video-based spatial reasoning tasks. To ensure data quality, answer balancing and visibility filtering is adopted for spatial QA samples. Camera motion annotations from CameraBench are manually verified, and semantic scene captions generated by Gemini-2.5-flash are manually inspected on a representative subset to confirm semantic accuracy and consistency with camera motion.

\subsection{Caption Generation Prompt}
\label{sec:caption-prompt}

Semantic scene captions are generated using Gemini-2.5-flash~\citep{comanici2025gemini} with a fixed prompt designed to strictly separate scene content from camera motion:

\begin{quote}

\texttt{Provide a concise description of the scene and objects visible in this video. Focus strictly on the environment and static/dynamic objects.
Do NOT describe the camera motion (ignore zooming, panning, or shakiness).}

\end{quote}

\subsection{Dataset Templates}
\label{sec:template}
We employ 25 different prompt templates to increase robustness and prompt diversity. All spatial narrative templates explicitly separate semantic content and camera motion using structured tags. Several example templates are shown below:

\begin{itemize}
  \item \texttt{What is the scene description and the camera behavior of the video? Separate the caption into $<$scene$>$ and $<$camera$>$.}
  \item \texttt{Describe what is happening in the video and how the camera moves. Use $<$scene$>$ for the content and $<$camera$>$ for the camera motion.}
  \item \texttt{Summarize the video by describing the visual scene and the camera dynamics, formatted as $<$scene$>$ and $<$camera$>$.}
  \item \texttt{Provide a high-level scene overview and a detailed camera movement description, using $<$scene$>$ and $<$camera$>$ tags.}
  \item \texttt{Explain the visual content of the video and the camera operation separately in $<$scene$>$ and $<$camera$>$.}
  \item \texttt{What does the video depict, and how is the camera positioned or moved? Answer using $<$scene$>$ and $<$camera$>$.}
  \item \texttt{Describe the main visual elements in the video and the camera’s motion or stability, separated into $<$scene$>$ and $<$camera$>$.}
  \item \texttt{What is shown in the video, and does the camera pan, tilt, zoom, or remain static? Use $<$scene$>$ and $<$camera$>$.}
  \item \texttt{Provide a structured description of the video’s content and camera behavior using $<$scene$>$ and $<$camera$>$ tags.}
\end{itemize}

\begin{figure}[h]
\centering
\includegraphics[width=0.98\linewidth]{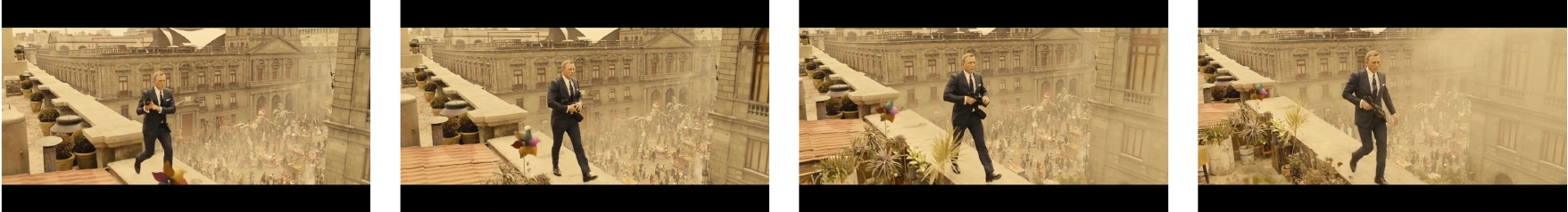}
    \caption{Video example from \datasetname.}
    \label{fig:exp1}
\end{figure}

\begin{figure}[h]
\centering
\includegraphics[width=0.98\linewidth]{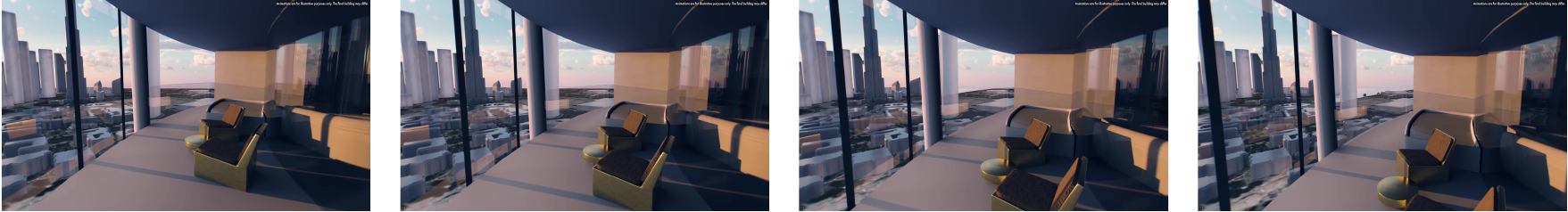}
    \caption{Video example from \datasetname.}
    \label{fig:exp2}
\end{figure}

\begin{figure}[H]
\centering
\includegraphics[width=0.98\linewidth]{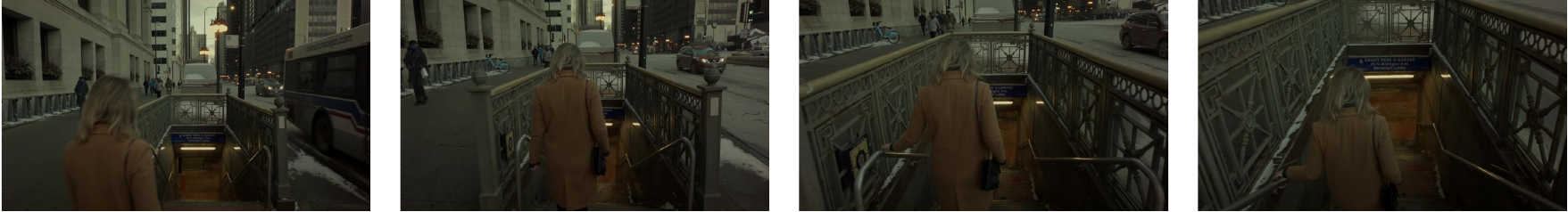}
    \caption{Video example from \datasetname.}
    \label{fig:exp3}
\end{figure}

\subsection{Dataset Examples}
We showcase several video examples and corresponding spatial narrative annotations from \datasetname~in Figure~\ref{fig:exp1} and Figure~\ref{fig:exp2}. The corresponding spatial narrative annotations are as followings:

\paragraph{Figure~\ref{fig:exp1} Spatial Narrative Annotations.} \textit{$<$scene$>$ A man in a dark suit, white shirt, and tie runs along a narrow stone ledge or rooftop, holding a pistol in his right hand. Potted plants and a colorful pinwheel are visible on the ledge and surrounding lower rooftops. Below him, a large, bustling street parade or festival unfolds, filled with numerous people, many in costumes (including large skeleton figures) and elaborate decorated floats or carts. Old, multi-story stone buildings with many windows and architectural details line the street and fill the background, with a particularly grand building featuring pillars prominent in the far background. The entire scene is enveloped in a general haze or dust, giving it a yellowish-brown tint. $<$camera$>$ The camera smoothly tracks backward from the left-front of the subject, initially tilting slightly downward as it follows the person from the front-side with minimal shaking.}

\paragraph{Figure~\ref{fig:exp2} Spatial Narrative Annotations.} \textit{$<$scene$>$ The video shows a modern, open-air balcony with a glass railing and a light-colored floor. Two gold chairs with dark cushions and a small gold table are placed on the balcony. In the background, there is a panoramic view of a city skyline with several tall buildings and a large body of water, possibly a sea or ocean, under a cloudy sky. $<$camera$>$ The camera dolly forward and pedestals up in a smooth, seamless motion, maintaining perfect steadiness throughout.}

\paragraph{Figure~\ref{fig:exp3} Spatial Narrative Annotations.} \textit{$<$scene$>$ A woman with blonde hair in a brown coat walks down a snow-dusted metal staircase into an underground entrance. Above the stairs, a blue sign reads "GRANT PARK N GARAGE, 25 N Michigan Ave, Randolph Lobby". On the street level, a large bus drives past on the right, and other pedestrians walk along a sidewalk on the left. Tall city buildings line the street, and a blue bicycle is parked against a building on the left. There is some snow visible on the ground and railings. $<$camera$>$ The camera smoothly dollies forward while tilting down, following the subject from behind with minimal shaking.}

\clearpage

\subsection{Training Recipe}
We adopt Qwen2.5-VL-3B-Instruct~\citep{bai2025qwen2} as the base model and perform supervised fine-tuning using the LLaMA-Factory~\citep{zheng2024llamafactory} codebase. Training is conducted for a single epoch with full-parameter fine-tuning on the LLM, while freezing the vision encoder and multi-modal projector. All the training and inference are conducted on 8 H200 GPU. We summarize the detail training recipes in Table~\ref{table:training_config}. Our model checkpoint, data, training and evaluation code will be released upon acceptance.

\begin{table}[t]
\centering
\caption{Training configuration for CaMo-3B.}
\label{table:training_config}
\resizebox{0.6\linewidth}{!}{
\begin{tabular}{l|l}
\toprule
\textbf{Hyperparameter} & \textbf{Setting} \\
\midrule
Base model & Qwen2.5-VL-3B-Instruct \\
Training method & Supervised Fine-tuning (SFT) \\
Finetuning type & Full-parameter \\
Frozen modules & Vision tower, multimodal projector \\
Optimizer & AdamW \\
Learning rate & $1.0 \times 10^{-5}$ \\
Scheduler & Cosine with warmup ratio 0.1 \\
Epochs & 1 \\
Batch size (per device) & 8 \\
Gradient accumulation & 2 \\
Precision & BF16 \\
Max sequence length & 4096 \\
Video FPS & 8 \\
Flash attention & FA2 \\
Deepspeed & ZeRO-2 \\
\bottomrule
\end{tabular}}
\end{table}

\section{Spatial Narrative Score Evaluation Details}

\subsection{Evaluation Settings}
\label{sec:evaluation_setting}
For Spatial Narrative Score (SNS) evaluation, each input video is uniformly partitioned into fixed-length temporal segments without overlap, with each segment consisting of 16 frames. The VLM generates a spatial narrative for each segment, which are then concatenated into a video-level spatial narrative used for downstream reasoning. We conduct SNS evaluation on VSI-Bench-tiny, which contains 198 multiple-choice questions (MCQ). This subset enables reproducible and cost-efficient evaluation, requiring approximately 30 minutes for LLM proxy answer generation per SNS run, compared to over 7 hours for the full benchmark, thereby facilitating rapid iteration and broader adoption by the community. For spatial narrative generation in SNS evaluation and camera motion captioning, we adopt the following prompt:

\begin{quote}

\texttt{Describe what is happening in the video and how the camera moves.\textbackslash n \\Use $<$scene$>$ for the content and $<$camera$>$ for the camera motion.}

\end{quote}

For spatial question answering, we follow existing evaluation codebase~\citep{li2025spatialladder} and adopt the following prompt:

\begin{quote}

\medskip
\textbf{Multiple-Choice Question.} \\
\texttt{Please answer with the option's letter from the given choices (e.g., A, B, etc.) directly.}

\medskip
\textbf{Numerical Question.} \\
\texttt{Please answer the question using a numerical value (e.g., 42 or 3.1) directly.}
\end{quote}

\subsection{Evaluation Prompt}
\label{sec:evaluation_prompt}
The proxy LLM is prompted to reason over the provided spatial narratives and answer the corresponding question. The proxy does not observe the video directly and relies solely on the narrative representation. We use Gemini-2.5-flash as the default proxy model with a fixed reasoning prompt and a thinking token budget of 1024. The prompt used for SNS evaluation is shown below, with the corresponding video spatial narrative, question, and options specified in the brackets replaced by the corresponding content.

\begin{quote}
\textbf{Instruction.} You are provided with multiple segments of dense 3D scene captions from a continuous video. Note that there may be multiple objects of the same category in the scene. Use the described camera motion to infer the spatial layout and answer the given question. You must base your answer on explicit reasoning and your best judgment.

\medskip
\textbf{Video Captions.} \texttt{\{video spatial narrative\}}

\medskip
\textbf{Question.} \texttt{\{question\}}

\medskip
\textbf{Options.} \texttt{\{options\}}

\medskip
\textbf{Final Instruction.} You must provide the final answer using the exact format:
\texttt{<answer>LETTER</answer>}. Example: \texttt{<think>your reasoning</think> <answer>A</answer>}
\end{quote}

\subsection{Ablation on Different Proxy LLMs}

\begin{table}[h]
\centering
\caption{Ablation on different proxy LLMs for SNS Evaluation on the same question and spatial narrative set.}
\label{tab:ablation_proxy_llms_row}
\resizebox{0.5\linewidth}{!}{
\begin{tabular}{l | c c c c c}
\toprule
\textbf{Proxy LLM} 
& Gemini-3-flash-preview
& Gemini-2.5-flash 
& GPT-5.1
& GPT-5-mini 
& GPT-5-nano \\
\midrule
\textbf{SNS} 
& 50.0 
& 51.0 
& 50.0
& 47.4 
& 48.0 \\
\bottomrule
\end{tabular}
}
\end{table}
To study the sensitivity of SNS to the choice of proxy LLM, we evaluate the spatial narratives score using a diverse collection of proxy LLMs. We focus on several cost-efficient flagship proxy LLMs~(Gemini-flash series, GPT-5 series) with strong textual spatial reasoning ability~\citep{guo2026llmspixelsbenchmarkingspatial}. These model selection and evaluation setting substantially reduces evaluation cost and makes SNS more accessible and friendly for public reproduction. As shown in Table~\ref{tab:ablation_proxy_llms_row}, the resulting SNS scores remain consistent across different proxies, with all models yielding comparable performance. This suggests that SNS primarily reflects the quality of the spatial narratives rather than being tightly coupled to a specific LLM.

\section{Experiment Details}
\label{sec:appendix_exp}
\subsection{Evaluated Benchmarks}
\label{sec:benchmark}
\begin{itemize}
    \item \textbf{VSI-Bench}~\citep{yang2025thinking}: VSI-Bench is a large-scale benchmark designed to evaluate visual-spatial intelligence in Multimodal Large Language Models (MLLMs). It contains more than 5,000 question–answer pairs derived from 288 real-world scenes collected from ScanNet~\cite{dai2017scannet}, ScanNet++~\cite{yeshwanth2023scannet++}, and ARKitScenes~\citep{baruch2021arkitscenes}. These videos span a wide variety of indoor environments across different geographic locations, enabling comprehensive assessment of spatial perception and reasoning.
    
    \item \textbf{SPBench-SI \& SPBench-MV}: SPBench-SI and SPBench-MV~\citep{li2025spatialladder} are evaluation benchmarks collected from the ScanNet validation split. SPBench-SI focuses on single-image spatial reasoning from individual viewpoints and includes four task categories: absolute distance, object size, relative distance, and relative direction, totaling 1,009 samples. SPBench-MV targets multi-view spatial reasoning, requiring models to jointly reason across multiple viewpoints. In addition to spatial relations, SPBench-MV introduces object counting tasks to assess multi-view object identification and enumeration, comprising 319 samples.
    
    \item \textbf{CV-Bench}~\citep{tong2024cambrian}: CV-Bench is proposed to address the shortcomings of existing vision-centric benchmarks and consists of 2,638 manually verified examples. It repurposes well-established vision datasets, including ADE20k~\citep{zhou2017ade20k}, COCO~\citep{lin2014coco}, and OMNI3D~\citep{brazil2023omni3d}, to evaluate MLLMs on core computer vision tasks. The benchmark measures 2D spatial understanding through spatial relations and object counting, and evaluates 3D reasoning via depth ordering and relative distance estimation.
    
    \item \textbf{SPAR-Bench}~\citep{zhang2025flatland}: SPAR-Bench is a comprehensive evaluation suite for systematically measuring spatial perception and reasoning capabilities in Vision–Language Models (VLMs). It includes 20 diverse spatial reasoning tasks covering single-view images, multi-view settings, and temporal video scenarios. The benchmark contains 7,207 manually validated question–answer pairs, ensuring high annotation accuracy and reliability.
    
    \item \textbf{ViewSpatial-Bench}~\citep{li2025viewspatial}: ViewSpatial-Bench is a large-scale benchmark comprising over 5,700 question–answer pairs across more than 1,000 3D scenes from the ScanNet~\citep{dai2017scannet} and MS-COCO~\citep{lin2014coco} validation datasets. It evaluates VLMs’ spatial localization and reasoning abilities from both egocentric and allocentric perspectives, targeting the crucial challenge of perspective understanding required for embodied agents and multi-agent interaction.
\end{itemize}

\subsection{Compared Baselines}
\begin{itemize}
    \item \textbf{GPT-4o}~\citep{hurst2024gpt}: GPT-4o is a multilingual and multi-modal generative pretrained transformer developed by OpenAI. The model supports unified processing and generation of text, images, and audio, demonstrating strong cross-modal reasoning capabilities across multiple languages.
    
    \item \textbf{Gemini-2.0-Flash}~\citep{team2024gemini}: Gemini 2.0 Flash is a multimodal language model tailored for agent-centric applications. It emphasizes computational efficiency, native tool use, multimodal content generation, and supports an extended context window of up to one million tokens. Compared to earlier Flash variants, it achieves higher output quality while maintaining similar inference speed.
    
    \item \textbf{InternVL-2.5-4B/8B}~\citep{chen2024internvl}: InternVL 2.5 is an advanced MLLM that extends InternVL 2.0 with improved training methodologies and higher-quality data. It achieves competitive results across a broad range of benchmarks, including reasoning, document understanding, and video comprehension, and performs on par with commercial systems such as GPT-4o and Claude-3.5-Sonnet.
    
    \item \textbf{Kimi-VL-A3B}~\citep{team2025kimi}: Kimi-VL-A3B is an efficient open-source Vision–Language Model based on a Mixture-of-Experts (MoE)~\citep{jiang2024mixtral} architecture. It provides strong multimodal reasoning, long-context understanding, and agent capabilities while activating only 2.8B parameters in the language decoder during inference.
    
    \item \textbf{LLaVA-Onevision-7B}~\citep{li2024llava-one}: LLaVA-OneVision-7B is an open-source MLLM that performs robustly across single-image, multi-image, and video-based tasks. The model exhibits effective cross-modal transfer, with notable improvements in video understanding emerging from training on image-centric tasks.
    
    \item \textbf{Qwen2.5-VL-3B\&7B}~\citep{bai2025qwen2}: Qwen2.5-VL is part of the Qwen family of open Vision–Language Models, featuring enhanced capabilities in visual recognition, object grounding, document analysis, and long-video understanding. Additionally, Qwen2.5-VL achieves strong performance on general video captioning task~\citep{wang2025vdc}, therefore we consider it as a strong representative model of general video captioning VLMs in our spatial narrative score evaluation.

    \item \textbf{SpatialLadder-3B}~\citep{li2025spatialladder}: SpatialLadder is a recent spatial understanding MLLM that leverages multi-stage training with strong spatial reasoning capability. It leverages supervised fine-tuning and GRPO-training on the collected SpatialLadder-26K, achieving strong performance on VSI-Bench and other spatial reasoning benchmarks.
    
    \item \textbf{SpaceR-3B\&7B}~\citep{ouyang2025spacer}: SpaceR is a VLM specifically developed for spatial reasoning using reinforcement learning with verifiable rewards. It introduces a map imagination GRPO that enables the model to infer spatial layouts during reasoning, achieving strong performance and substantially outperforming GPT-4o on VSI-Bench.
    
    \item \textbf{VILASR-7B}~\citep{wu2025reinforcing}: VILASR proposes a "drawing to reason in space" paradigm, allowing VLMs to perform spatial reasoning via simple drawing operations such as bounding box annotation and auxiliary line construction. The model is trained using a three-stage pipeline consisting of synthetic data pretraining, reflective rejection sampling, and reinforcement learning. It achieves superior performance across a wide range of spatial reasoning benchmarks.
    
    \item \textbf{Video-R1}~\citep{feng2025video}: Video-R1 extends the R1 reasoning framework to video understanding in MLLMs, following the reinforcement learning methodology of DeepSeek-R1~\citep{guo2025deepseek}. It adopts the T-GRPO algorithm to better exploit temporal information and is trained on both image- and video-based reasoning data. Video-R1-7B demonstrates strong performance on video reasoning benchmarks, surpassing GPT-4o on VSI-Bench and performing robustly on general video tasks.
    
    \item \textbf{Spatial-MLLM-4B}~\citep{wu2025spatial}: Spatial-MLLM is a unified framework for visual-based spatial reasoning that employs a dual-encoder architecture. It combines a pretrained 2D visual encoder for semantic representation with a spatial encoder initialized from a visual geometry foundation model~\citep{wang2025vggt} to capture 3D structural information, and achieves SOTA results across multiple spatial understanding and reasoning benchmarks.
\end{itemize}

\begin{table}[ht]
\centering
\small
\caption{\textbf{Direct QA vs. Spatial Narrative Score (SNS) on VSI-Bench.}
Each cell shows \emph{Direct / SNS} with the gap \emph{(SNS$-$Direct)}.}
\label{tab:direct_vs_sns_allcats}
\resizebox{\linewidth}{!}{
\begin{tabular}{lccccc}
\toprule
\textbf{Model} & \textbf{Rel. Dir.} & \textbf{Rel. Dist.} & \textbf{Appr. Order} & \textbf{Route Plan.} & \textbf{Overall} \\
\midrule
SpatialLadder &
52.0 / 30.0 \;(\textcolor{DarkRed}{-22.0}) &
56.0 / 32.0 \;(\textcolor{DarkRed}{-24.0}) &
42.9 / 38.8 \;(\textcolor{DarkRed}{-4.1}) &
33.3 / 30.6 \;(\textcolor{DarkRed}{-2.7}) &
46.0 / 32.8 \;(\textcolor{DarkRed}{-13.2}) \\
\methodname &
47.4 / 40.0 \;(\textcolor{DarkRed}{-7.4}) &
36.0 / 46.0 \;(\textcolor{ForestGreen}{+10.0}) &
63.3 / 69.4 \;(\textcolor{ForestGreen}{+6.1}) &
20.4 / 49.0 \;(\textcolor{ForestGreen}{+28.6}) &
41.8 / 51.0 \;(\textcolor{ForestGreen}{+9.2}) \\
\bottomrule
\end{tabular}
}
\end{table}

\subsection{Detailed direct QA vs. SNS Results}
Table~\ref{tab:direct_vs_sns_allcats} contrasts Direct QA accuracy with Spatial Narrative Score (SNS). A clear divergence emerges between the two evaluation interfaces. SpatialLadder achieves higher Direct QA accuracy in most categories, yet exhibits consistent performance drops under SNS, particularly in \emph{Rel. Dir.} and \emph{Rel. Dist.}, indicating limited transfer of its spatial reasoning beyond answer prediction. In contrast, \methodname~shows a markedly different trend: while its Direct QA accuracy is comparable or lower in some categories, SNS performance improves substantially, yielding positive gaps in \emph{Rel. Dist.}, \emph{Appr. Order}, \emph{Route Plan.}, and \emph{Overall} results. These results suggest that \methodname~learns spatial representations that are more robustly externalized through narrative generation, enabling effective downstream reasoning. Overall, the table highlights that strong Direct QA accuracy does not necessarily translate to coherent spatial narratives, whereas improvements under SNS reflect more transferable spatial understanding.

\subsection{Spatial Narrative Score Evaluation on the Full VSI-Bench}

\begin{table}[h]
\centering
\small
\caption{\textbf{Comparison of SNS Evaluation results on the full VSI-Bench.} Improvements are relative to the Qwen2.5-VL-3B. \textbf{Bold} indicates best performance, while \underline{underlined} represents second-best.}
\label{tab:vsi-full-comparison-inline}
\resizebox{\textwidth}{!}{
\begin{tabular}{lcccccc}
\toprule
\textbf{Model} & \textbf{Training} & \textbf{Rel. Dir.} & \textbf{Rel. Dist.} & \textbf{Appr. Order} & \textbf{Route Plan.} & \textbf{Overall} \\
\midrule
Qwen2.5-VL-3B~\citep{bai2025qwen2} & SFT 
& 31.2 
& 42.0 
& 56.1 
& 41.2 
& 41.2 \\
SpatialLadder-3B~\citep{li2025spatialladder} & GRPO 
& 31.9 (\textcolor{ForestGreen}{+0.7}) 
& 38.7 (\textcolor{DarkRed}{-3.3}) 
& 46.3 (\textcolor{DarkRed}{-9.8}) 
& 35.6 (\textcolor{DarkRed}{-5.6}) 
& 37.7 (\textcolor{DarkRed}{-3.5}) \\
\midrule
\rowcolor{gray!10}
\textbf{\methodname} & SFT 
& \textbf{33.7} (\textcolor{ForestGreen}{+2.5}) 
& \textbf{46.3} (\textcolor{ForestGreen}{+4.3}) 
& \textbf{58.7} (\textcolor{ForestGreen}{+2.6}) 
& \textbf{43.3} (\textcolor{ForestGreen}{+2.1}) 
& \textbf{44.3} (\textcolor{ForestGreen}{+3.1}) \\
\bottomrule
\end{tabular}
}
\end{table}

We present detailed evaluation results on the full VSI-Bench MCQ in Table~\ref{tab:vsi-full-comparison-inline}, which we compare the spatial narrative score (SNS) of multiple models, including the base model Qwen2.5-VL-3B, GRPO-based spatial understanding model SpatialLadder-3B~\citep{li2025spatialladder}, and our model \methodname. Compared with the Qwen2.5-VL-3B baseline, \methodname~consistently improves performance across all four SNS question categories, achieving the best overall score of 44.3 with a +3.1 absolute gain. In contrast, SpatialLadder-3B, despite using GRPO-based training, underperforms the baseline on most categories, indicating that optimization alone is insufficient without effective spatial representations. The improvements of \methodname~on relative distance, appearance order, and route planning demonstrate its stronger ability to generate spatial narratives with better quality and more accurate spatial information.

\subsection{VSI-Bench Results}
Table \ref{tab:vsibench} reports comprehensive results on VSI-Bench, which evaluates fine-grained spatial reasoning and long-horizon navigation across numerical and multiple-choice questions. While larger proprietary and 7B-scale spatial models achieve strong performance on individual subtasks, \methodname~significantly improves over the Qwen2.5-VL-3B baseline with a +14.6 increase in average accuracy. The gains are most pronounced in object size (+41.9), room size (+11.7), relative distance (+12.4), and approach order (+21.4), demonstrating stronger geometric understanding and sequential spatial reasoning. Although minor regressions appear on absolute distance and route planning, \methodname~achieves competitive overall performance against larger 7B models, highlighting the effectiveness of the proposed spatial modeling in capturing both metric and narrative spatial structures within a compact model.

\subsection{SP-Bench Results}
Table \ref{tab:spbench-si} reports results on SPBench-SI, evaluating both numerical and multiple-choice spatial reasoning. Overall, proprietary models such as GPT-4o and Gemini-2.0-Flash exhibit competitive performance, particularly on relative distance questions, but remain limited on fine-grained numerical estimation and relational direction understanding. Notably, \methodname~achieves a substantial performance gain over all Qwen2.5-VL-3B–based spatial models, improving the average score by +24.1 points. The largest improvements are observed on object size estimation (+53.2) and relative direction reasoning (+35.9), indicating that \methodname~more effectively captures both metric and relational spatial cues. Despite using a smaller backbone, \methodname~rivals or surpasses several 7B-scale spatial models, highlighting the effectiveness of its spatial reasoning design rather than parameter count. These results demonstrate that targeted spatial modeling can significantly enhance spatial intelligence, especially for challenging relational and directional tasks.

Table \ref{tab:spbench-mv} presents results on SPBench-MV, which evaluates multi-view spatial reasoning across object counting, metric estimation, and relational understanding. While proprietary and large open-source models achieve strong performance on individual subtasks, their results remain unbalanced, particularly on relational direction and object size estimation. In contrast, \methodname~consistently outperforms all baselines, achieving the highest average score of 65.8 and setting new best results on object counting, object size, relative distance, and relative direction. Notably, \methodname~improves upon the strongest Qwen2.5-VL-3B–based baseline by +29.2 points on average, with especially large gains in object counting (+55.4) and object size (+58.0), demonstrating its superior ability to integrate multi-view geometric cues. These results highlight that effective multi-view spatial modeling, rather than increased model scale alone, is critical for robust spatial reasoning.

\clearpage
\begin{table}[t]
\centering
\small
\caption{\textbf{Evaluation results on VSI-Bench.}}
\label{tab:vsibench}
\resizebox{0.9\textwidth}{!}{
\begin{tabular}{lccccccccc}
\toprule
\multirow{2}{*}{\textbf{Model}} & \multicolumn{4}{c}{\textbf{Numerical Question}} & \multicolumn{4}{c}{\textbf{Multiple-choice Question}} & \multirow{2}{*}{\textbf{Avg.}} \\
\cmidrule(lr){2-5} \cmidrule(lr){6-9}
& Obj. Cnt & Abs. Dist. & Obj. Size & Room Size & Rel. Dist. & Rel. Dir. & Route Plan. & Appr. Order & \\
\midrule
\rowcolor{gray!10} 
\multicolumn{10}{l}{\textit{\textbf{Proprietary Models}}} \\
GPT-4o & 46.2 & 5.3 & 43.8 & 38.2 & 37.0 & 41.3 & 31.5 & 28.5 & 34.0 \\
Gemini-2.0-Flash & 56.2 & 30.9 & \textbf{66.7} & 31.8 & \textbf{51.3} & \underline{46.3} & 24.5 & 55.1 & 45.4 \\
\midrule
\rowcolor{gray!10} 
\multicolumn{10}{l}{\textit{\textbf{Open-Source Models}}} \\
InternVL-2.5-4B & 45.0 & 15.5 & 37.5 & 24.6 & 37.2 & 41.5 & 31.4 & 26.2 & 32.6 \\
InternVL-2.5-8B & 50.6 & 31.3 & 40.2 & 39.3 & 45.1 & 41.4 & 29.4 & 43.9 & 40.2 \\
Kimi-VL-A3B  & 41.3 & 30.4 & 42.1 & 13.2 & 26.3 & 32.6 & 32.0 & 11.2 & 28.7 \\
LLaVA-OneVision-7B & 46.1 & 26.2 & 36.3 & 29.5 & 30.8 & 37.2 & \underline{35.1} & 21.8 & 33.1 \\
\midrule
\rowcolor{gray!10} 
\multicolumn{10}{l}{\textit{\textbf{Qwen2.5-VL-7B Based Spatial Models}}} \\
Qwen2.5-VL-7B & 43.5 & 15.1 & 48.5 & \underline{41.1} & 36.3 & 40.1 & 28.4 & 33.7 & 35.8 \\
SpaceR-7B & 63.2 & 30.0 & 60.3 & 37.6 & 39.7 & 45.6 & 31.4 & 48.2 & 44.5 \\
VILASR-7B & 63.5 & \underline{34.4} & 60.6 & 30.9 & \underline{48.9} & 45.2 & 30.4 & 49.2 & 45.5 \\
Video-R1 & 34.0 & 23.0 & 41.6 & 36.7 & 36.8 & 34.7 & 31.4 & 28.8 & 33.4 \\
\midrule
\rowcolor{gray!10} 
\multicolumn{10}{l}{\textit{\textbf{Qwen2.5-VL-3B Based Spatial Models}}} \\
Qwen2.5-VL-3B & 32.9 & 22.1 & 17.3 & 31.5 & 32.8 & 44.2 & 26.3 & 28.5 & 29.4 \\
Spatial-MLLM-4B & \textbf{65.6} & \textbf{35.5} & \underline{64.2} & 40.6 & 41.3 & \textbf{47.9} & 34.0 & \underline{49.2} & \textbf{47.3} \\
\rowcolor{gray!10}
\textbf{\methodname} & 64.8 & 18.6 & 59.2 & \textbf{43.2} & 45.2 & 45.3 & 25.8 & \textbf{49.9} & 44.0 \\
\textcolor{ForestGreen}{\textit{\textbf{Improvement}}} & \textcolor{ForestGreen}{+31.9} & \textcolor{DarkRed}{-3.5} & \textcolor{ForestGreen}{+41.9} & \textcolor{ForestGreen}{+11.7} & \textcolor{ForestGreen}{+12.4} & \textcolor{ForestGreen}{+1.1} & \textcolor{DarkRed}{-0.5} & \textcolor{ForestGreen}{+21.4} & \textcolor{ForestGreen}{+14.6} \\
\bottomrule
\end{tabular}
}
\end{table}
\begin{table}[H]
\centering
\small
\caption{\textbf{Evaluation results on SPBench-SI.}}
\label{tab:spbench-si}
\resizebox{0.5\textwidth}{!}{
\begin{tabular}{lccccc}
\toprule
\multirow{2}{*}{\textbf{Model}} & \multicolumn{2}{c}{\textbf{Numerical Question}} & \multicolumn{2}{c}{\textbf{Multiple-choice Question}} & \multirow{2}{*}{\textbf{Avg.}} \\
\cmidrule(lr){2-3} \cmidrule(lr){4-5}
& Abs. Dist. & Obj. Size & Rel. Dist. & Rel. Dir. & \\
\midrule
\rowcolor{gray!10} 
\multicolumn{6}{l}{\textit{\textbf{Proprietary Models}}} \\
GPT-4o & 19.7 & 29.0 & 81.3 & 39.2 & 42.4 \\
Gemini-2.0-Flash & 33.1 & 64.9 & 81.3 & 39.5 & 54.7 \\
\midrule
\rowcolor{gray!10} 
\multicolumn{6}{l}{\textit{\textbf{Open-Source Models}}} \\
InternVL-2.5-4B & 27.3 & 36.2 & 73.6 & 33.0 & 42.5 \\
InternVL-2.5-8B & 15.6 & 40.8 & 76.9 & 35.6 & 42.3 \\
Kimi-VL-A3B  & 11.3 & 40.2 & 62.6 & 27.1 & 35.3 \\
LLaVA-OneVision-7B & 23.6 & 27.2 & 54.9 & 27.1 & 33.2 \\
\midrule
\rowcolor{gray!10} 
\multicolumn{6}{l}{\textit{\textbf{Qwen2.5-VL-7B Based Spatial Models}}} \\
Qwen2.5-VL-7B & 27.7 & 45.0 & \textbf{83.5} & 37.6 & 48.4 \\
SpaceR-7B & 8.4 & 62.9 & 80.2 & 42.8 & 48.6 \\
VILASR-7B & 10.3 & 63.0 & 81.3 & \underline{46.1} & 50.2 \\
Video-R1 & 5.1 & 50.3 & \underline{82.4} & 41.5 & 44.8 \\
\midrule
\rowcolor{gray!10} 
\multicolumn{6}{l}{\textit{\textbf{Qwen2.5-VL-3B Based Spatial Models}}} \\
Qwen2.5-VL-3B & 30.9 & 17.8 & 75.8 & 36.6 & 40.3 \\
Spatial-MLLM-4B & 16.4 & 59.7 & 69.2 & 29.4 & 43.7 \\
\rowcolor{gray!10}
\textbf{\methodname} & \textbf{31.9} & \textbf{71.0} & 82.4 & \textbf{72.5} & \textbf{64.4} \\
\textcolor{ForestGreen}{\textit{\textbf{Improvement}}} & \textcolor{ForestGreen}{+1.0} & \textcolor{ForestGreen}{+53.2} & \textcolor{ForestGreen}{+6.6} & \textcolor{ForestGreen}{+35.9} & \textcolor{ForestGreen}{+24.1} \\
\bottomrule
\end{tabular}
}
\end{table}
\begin{table}[H]
\centering
\small
\caption{\textbf{Evaluation results on SPBench-MV.}}
\label{tab:spbench-mv}
\resizebox{0.6\textwidth}{!}{
\begin{tabular}{lcccccc}
\toprule
\multirow{2}{*}{\textbf{Model}} & \multicolumn{3}{c}{\textbf{Numerical Question}} & \multicolumn{2}{c}{\textbf{Multiple-choice Question}} & \multirow{2}{*}{\textbf{Avg.}} \\
\cmidrule(lr){2-4} \cmidrule(lr){5-6}
& Obj. Cnt & Abs. Dist. & Obj. Size & Rel. Dist. & Rel. Dir. & \\
\midrule
\rowcolor{gray!10} 
\multicolumn{7}{l}{\textit{\textbf{Proprietary Models}}} \\
GPT-4o & 66.3 & 12.0 & 43.8 & 82.4 & 36.4 & 48.2 \\
Gemini-2.0-Flash & 49.9 & \textbf{40.7} & 65.1 & 76.5 & 25.0 & 51.4 \\
\midrule
\rowcolor{gray!10} 
\multicolumn{7}{l}{\textit{\textbf{Open-Source Models}}} \\
InternVL-2.5-4B & 65.1 & 24.0 & 23.9 & 82.4 & 20.5 & 43.2 \\
InternVL-2.5-8B & 50.0 & 25.0 & 37.0 & 88.2 & 6.8 & 41.4 \\
Kimi-VL-A3B & 13.7 & 23.3 & 33.0 & 76.5 & \underline{38.6} & 37.0 \\
LLaVA-OneVision-7B & 21.1 & 21.3 & 19.2 & 76.5 & 22.7 & 32.2 \\
\midrule
\rowcolor{gray!10} 
\multicolumn{7}{l}{\textit{\textbf{Qwen2.5-VL-7B Based Spatial Models}}} \\
Qwen2.5-VL-7B & 46.1 & 11.0 & 35.4 & 88.2 & 11.4 & 37.3 \\
SpaceR-7B & \underline{90.1} & 33.7 & 65.1 & 82.4 & 25.0 & 59.4 \\
VILASR-7B & 65.3 & 34.7 & \underline{68.7} & 88.2 & 29.5 & 61.8 \\
Video-R1 & 33.9 & 18.0 & 45.6 & 76.5 & 29.5 & 40.7 \\
\midrule
\rowcolor{gray!10} 
\multicolumn{7}{l}{\textit{\textbf{Qwen2.5-VL-3B Based Spatial Models}}} \\
Qwen2.5-VL-3B & 36.7 & 14.7 & 14.9 & 88.2 & 18.2 & 36.6 \\
Spatial-MLLM-4B & 88.9 & 31.0 & 71.2 & \underline{88.2} & 29.5 & \underline{61.8} \\
\rowcolor{gray!10}
\textbf{\methodname} & \textbf{92.1} & \underline{18.7} & \textbf{72.9} & \textbf{100.0} & \textbf{45.5} & \textbf{65.8} \\
\rowcolor{gray!10}
\textcolor{ForestGreen}{\textit{\textbf{Improvement}}} & \textcolor{ForestGreen}{+55.4} & \textcolor{ForestGreen}{+4.0} & \textcolor{ForestGreen}{+58.0} & \textcolor{ForestGreen}{+11.8} & \textcolor{ForestGreen}{+27.3}  & \textcolor{ForestGreen}{+29.2}\\
\bottomrule
\end{tabular}
}
\end{table}

\begin{table}[h]
\centering
\small
\caption{\textbf{Evaluation results on CV-Bench.} For each metric, \textbf{bold} numbers indicate the best performance, while \underline{underlined} numbers represent the second-best performance.}
\label{tab:cv-bench}
\resizebox{0.62\textwidth}{!}{
\begin{tabular}{lccccc}
\toprule
\multirow{2}{*}{\textbf{Model}} & \multicolumn{3}{c}{\textbf{2D}} & \multicolumn{1}{c}{\textbf{3D}} & \multirow{2}{*}{\textbf{Overall}} \\
\cmidrule(lr){2-4}
& ADE20K & COCO & Avg. & Omni3D & \\
\midrule
GPT-4o & 65.1 & 73.8 & 69.4 & \underline{81.3} & \underline{75.4} \\
InternVL-2.5-4B & \underline{68.6} & \underline{78.5} & \underline{73.5} & 75.1 & 74.4 \\
Kimi-VL-A3B & 41.9 & 42.4 & 41.9 & 54.7 & 48.3 \\
LLaVA-OneVision-7B & 50.6 & 55.8 & 53.2 & 63.5 & 58.3 \\
Qwen2.5-VL-7B & \textbf{69.5} & \textbf{80.5} & \textbf{75.0} & \textbf{83.1} & \textbf{79.0} \\
\midrule
\rowcolor{gray!10}
Qwen2.5-VL-3B & 63.2 & 75.0 & 69.1 & 72.2 & 70.6 \\
\rowcolor{gray!10}
\textbf{\methodname} & \textbf{68.4} & \textbf{78.5} & \textbf{73.5} & \textbf{79.2} & \textbf{76.3} \\
\rowcolor{gray!10}
\textcolor{ForestGreen}{\textit{\textbf{Improvement}}} & \textcolor{ForestGreen}{+5.2} & \textcolor{ForestGreen}{+3.5} & \textcolor{ForestGreen}{+4.4} & \textcolor{ForestGreen}{+7.0} & \textcolor{ForestGreen}{+5.7} \\
\bottomrule
\end{tabular}
}
\end{table}
\subsection{CV-Bench Results}
Table \ref{tab:cv-bench} evaluates general vision understanding on CV-Bench across 2D and 3D tasks. While Qwen2.5-VL-7B achieves the strongest overall performance among all models, \methodname~ substantially improves over its 3B-scale backbone, yielding consistent gains across both 2D (ADE20K and COCO) and 3D (Omni3D) benchmarks. In particular, \methodname~improves the overall score by +5.7 points and achieves a notable +7.0 gain on the 3D Omni3D benchmark, indicating enhanced 3D-aware visual understanding. These results demonstrate that the proposed spatial reasoning enhancements do not compromise general visual perception, but instead strengthen both 2D and 3D understanding despite using a smaller backbone model.

\vspace{20pt}

\begin{table}[h]
\centering
\small
\caption{\textbf{Evaluation results on ViewSpatial-Bench.} For each metric, \textbf{bold} numbers indicate the best performance, while \underline{underlined} numbers represent the second-best performance.}
\label{tab:vs-bench}
\resizebox{0.9\textwidth}{!}{
\begin{tabular}{lcccccccc}
\toprule
\multirow{2}{*}{\textbf{Model}} & \multicolumn{3}{c}{\textbf{Camera Perspective}} & \multicolumn{4}{c}{\textbf{Person Perspective}} & \multirow{2}{*}{\textbf{Overall}} \\
\cmidrule(lr){2-4} \cmidrule(lr){5-8}
& Rel. Dir. & Obj. Ori. & Avg. & Obj. Ori. & Rel. Dir. & Sce. Sim. & Avg. & \\
\midrule
GPT-4o & 41.5 & 19.6 & 33.7 & 41.2 & 32.8 & 21.9 & 31.5 & 32.6 \\
InternVL-2.5-4B & 37.1 & \underline{31.8} & 40.8 & 43.6 & \underline{37.1} & 26.1 & 35.1 & 37.9 \\
Kimi-VL-A3B & 26.9 & 22.1 & 25.1 & \underline{63.1} & \textbf{43.9} & 20.3 & 41.5 & 33.6 \\
LLaVA-OneVision-7B & 29.8 & 26.1 & 28.5 & 22.4 & 31.0 & 26.9 & 26.5 & 27.5 \\
Qwen2.5-VL-7B & \underline{47.8} & 30.9 & \textbf{41.8} & 41.6 & 35.4 & \underline{26.9} & 39.8 & \underline{37.9} \\
\midrule
\rowcolor{gray!10}
Qwen2.5-VL-3B & 43.5 & \textbf{32.5} & 39.5 & 40.0 & 29.9 & 26.3 & 32.0 & 35.6 \\
\rowcolor{gray!10}
\textbf{\methodname} & 47.5 & 30.9 & \underline{41.5} & \textbf{53.9} & 31.5 & \textbf{35.1} & \textbf{40.4} & \textbf{41.0} \\
\rowcolor{gray!10}
\textcolor{ForestGreen}{\textit{\textbf{Improvement}}} & \textcolor{ForestGreen}{+4.0} & \textcolor{DarkRed}{-1.6} & \textcolor{ForestGreen}{+2.0} & \textcolor{ForestGreen}{+13.9} & \textcolor{ForestGreen}{+1.6} & \textcolor{ForestGreen}{+8.8} & \textcolor{ForestGreen}{+8.4} & \textcolor{ForestGreen}{+5.4}\\
\bottomrule
\end{tabular}
}
\end{table}

\subsection{View-Spatial Bench Results}
Table \ref{tab:vs-bench} reports results on ViewSpatial-Bench, evaluating perspective-aware spatial reasoning from both camera and person viewpoints. While existing models show uneven performance across perspectives, \methodname~achieves the best overall score of 41.0, outperforming its Qwen2.5-VL-3B backbone by +5.4 points. The gains are particularly pronounced under the person perspective, with substantial improvements in object orientation (+13.9) and scene similarity (+8.8), indicating stronger viewpoint transformation and embodied spatial understanding. Although a minor drop is observed in camera-based object orientation, \methodname~maintains competitive performance and achieves a higher camera-perspective average. These results demonstrate that the proposed method effectively enhances cross-perspective spatial reasoning, especially for person-centric understanding, without sacrificing camera-based performance.

\clearpage

\section{More Analysis of Spatial Narrative Generation}
\label{sec:wordcloud}
\begin{figure}[h]
  \centering
  \begin{minipage}{0.32\textwidth}
    \centering
    \includegraphics[width=\linewidth,height=100pt]{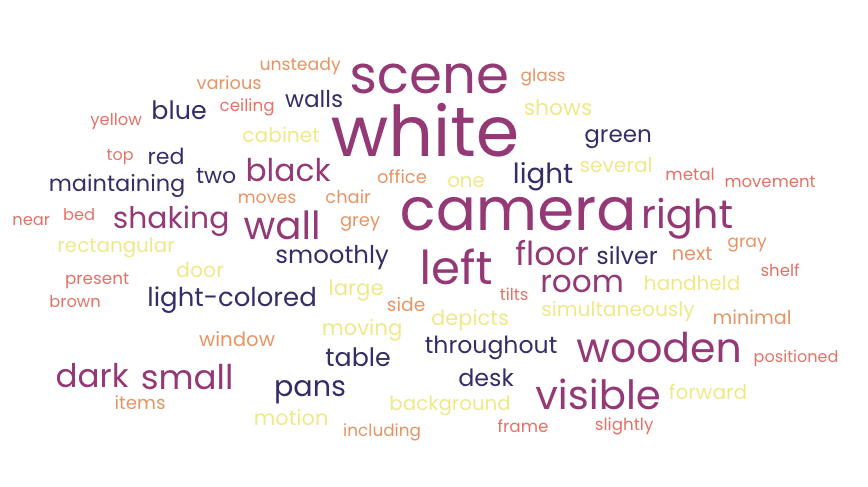}
    \captionof{figure}{Word cloud of \methodname.}
  \label{fig:wc-camo}
  \end{minipage}\hfill
  \begin{minipage}{0.32\textwidth}
    \centering
    \includegraphics[width=\linewidth,height=100pt]{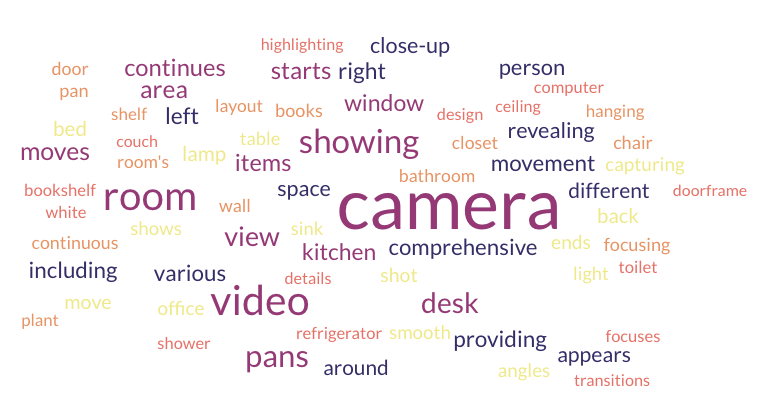}
    \captionof{figure}{Word cloud of SpatialLadder-3B.}
  \label{fig:wc-spld}
  \end{minipage}\hfill
  \begin{minipage}{0.32\textwidth}
    \centering
    \includegraphics[width=\linewidth,height=100pt]{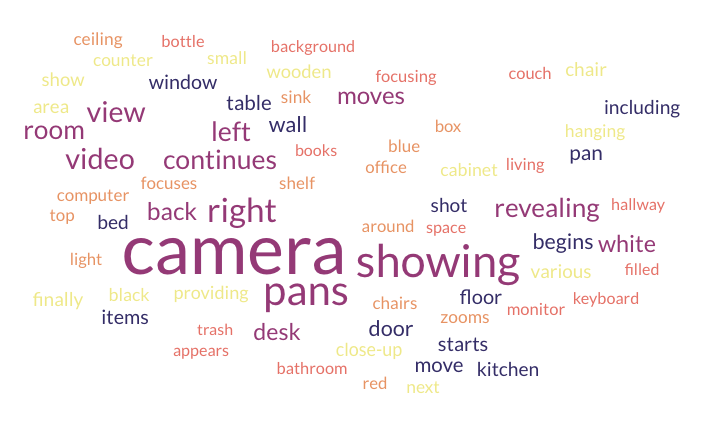}
    \captionof{figure}{Word cloud of Qwen2.5-VL-3B.}
  \label{fig:wc-qwen}
  \end{minipage}
\end{figure}

We present a word cloud analysis of the generated spatial narratives in Figures~\ref{fig:wc-camo}, \ref{fig:wc-spld}, and \ref{fig:wc-qwen} to qualitatively examine differences in language usage across models. Compared to the base model Qwen and our CaMo-3B, SpatialLadder (SPLD) exhibits a noticeably lower frequency of directional and egomotion-related terms, such as \textit{left}, \textit{right}, and camera motion descriptors. In contrast, narratives generated by CaMo-3B contain a richer set of directional words~(left, right) and motion-related expressions~(shaking, unsteady), indicating more explicit externalization of spatial relationships and camera movement. This observation aligns with our SNS results, suggesting that improved spatial narrative generation is associated with more frequent and explicit encoding of camera motion cues, rather than relying solely on object-centric or semantic descriptions.

\section{Case Study}
\label{sec:casestudy}
We perform detailed case study of \methodname~on a variety of videos for spatial narrative generation. As shown in the following, \methodname~is capable of generating detailed scene semantic and accurate camera motion captions for the spatial narrative evaluation.

\vspace{20pt}

\begin{figure}[h]
\centering
\includegraphics[width=0.98\linewidth]{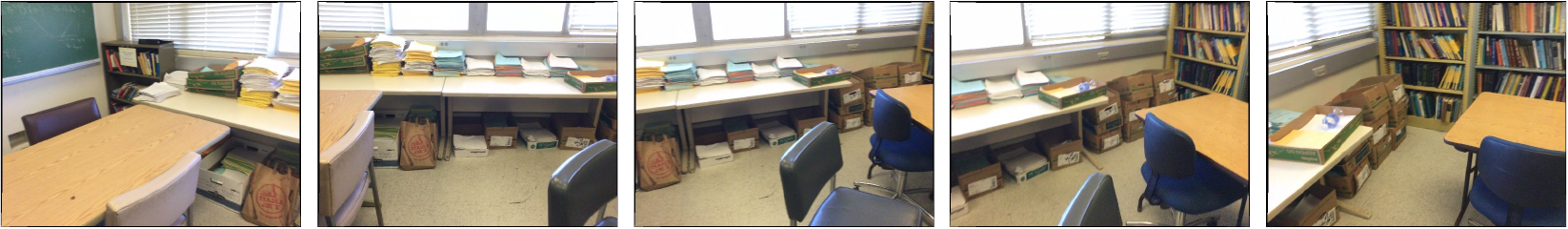}
    \caption{Spatial narrative video sample from CaMo.}
    \label{fig:case1}
\end{figure}

\paragraph{Figure~\ref{fig:case1}  \methodname~Generated Spatial Narrative.}  \textit{$<$scene$>$ The scene depicts an office or classroom space with a large window covered by blinds, allowing some light to filter through. A green chalkboard is visible on the left wall. On the left side of the room, there's a light-colored counter or desk holding various items including more stacks of papers, a dark green box, and several small boxes. Below this counter are multiple shelves filled with cardboard boxes of different sizes and contents. In the foreground, in the bottom right corner, a blue chair is positioned in front of a light wooden desk. Another empty desk is visible further back on the left. The back wall features two large bookshelves filled with books and files. The left bookshelf is partially visible in the background, while the right shelf displays numerous books neatly arranged in rows. $<$camera$>$ The camera smoothly pans right while simultaneously moving right, maintaining minimal shaking throughout the motion.}

\clearpage

\begin{figure}[h]
\centering
\includegraphics[width=0.98\linewidth]{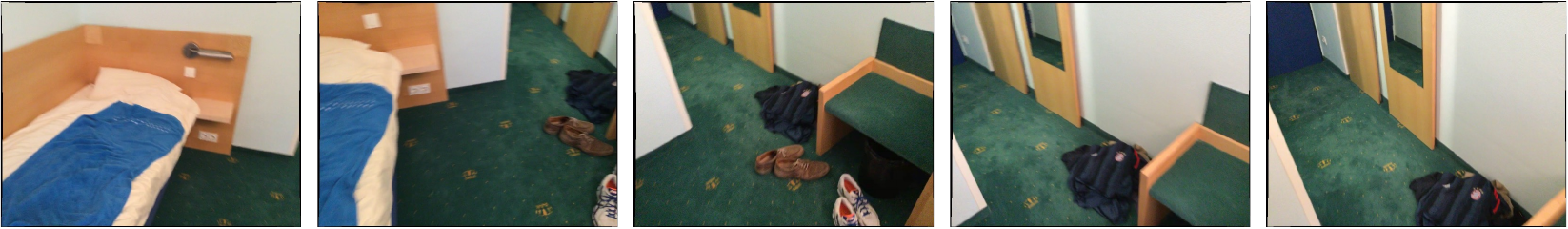}
    \caption{Spatial narrative video sample from CaMo.}
    \label{fig:case2}
\end{figure}

\begin{figure}[h]
\centering
\includegraphics[width=0.98\linewidth]{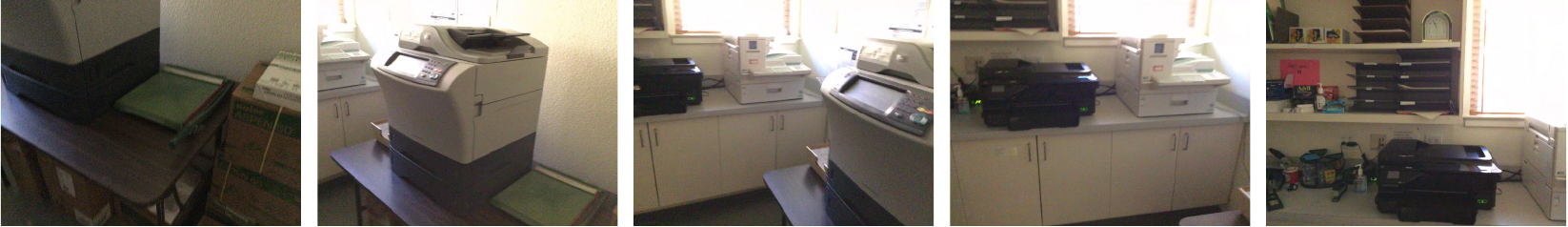}
    \caption{Spatial narrative video sample from CaMo.}
    \label{fig:case3}
\end{figure}

\begin{figure}[h]
\centering
\includegraphics[width=0.98\linewidth]{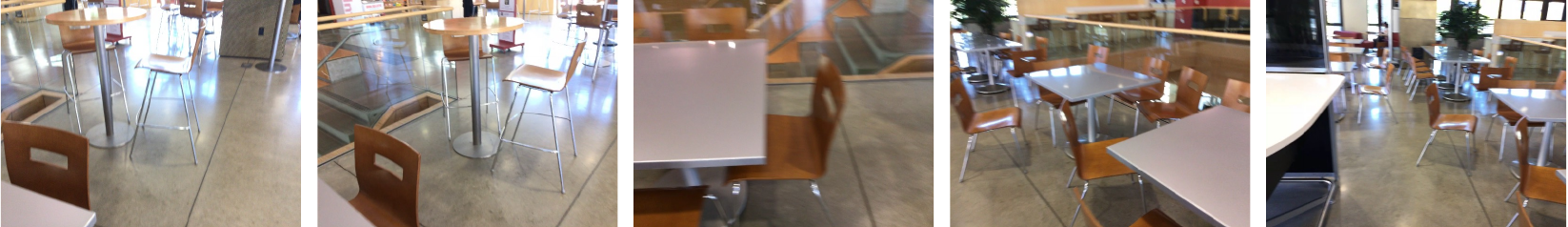}
    \caption{Spatial narrative video sample from CaMo.}
    \label{fig:case4}
\end{figure}

\vspace{10pt}

\paragraph{Figure~\ref{fig:case2} \methodname~Generated Spatial Narrative.}  \textit{$<$scene$>$ The scene depicts an indoor room with a bed featuring a wooden headboard, covered with a blue, white, and grey striped duvet and a white pillowcase. A green carpet covers the floor, adorned with a pattern of yellow crowns. Several pairs of shoes (including brown boots, white sneakers with blue laces, and brown loafers) are scattered on the carpet. A dark backpack rests next to the shoes. A wooden door with a silver handle and a peephole is visible, along with a light-colored wall and a section of blue wall. A rectangular mirror with a decorative frame hangs on the wall. $<$camera$>$ The camera pans right while tilting down from the headboard of the bed to the floor, then quickly moves forward to look at the backpack, all with an unsteady motion and noticeable shaking.}

\vspace{10pt}

\paragraph{Figure~\ref{fig:case3} \methodname~Generated Spatial Narrative.} \textit{$<$scene$>$ The scene depicts an office or study room with light-colored walls. On a large, light-colored desk, there's a black and gray laser printer, and next to it, a white and gray laser printer. A stack of brown cardboard boxes is also on the desk. In the background, there are light-colored storage cabinets with metal handles. Above the cabinets, a shelf holds binders, books, a red box of Advil, a blue container, and a small green clock. To the right, a window with blinds allows light to enter the room. $<$camera$>$ The handheld camera initially focuses on the printer before panning left and tilting up to reveal the entire scene of the office. It then pans right halfway through and tilts down at the end, maintaining a slightly unsteady movement with some shaking.}

\vspace{10pt}

\paragraph{Figure~\ref{fig:case4} \methodname~Generated Spatial Narrative.} \textit{$<$scene$>$ The scene depicts an indoor dining area with a modern aesthetic. The floor features light-colored, square-patterned tiles. Several glass-top tables with metallic bases are visible, some with wooden tops. Matching chairs with metallic frames and wooden seats and backs are also present. A whiteboard with a red frame stands in the foreground on a wooden surface. In the background, other dining areas and what appear to be kitchen or service counters are visible through large glass partitions. $<$camera$>$ The handheld camera smoothly pans left while simultaneously moving right, creating a slightly unsteady motion with some shaking.}

\clearpage

\begin{figure}[h]
\centering
\includegraphics[width=0.98\linewidth]{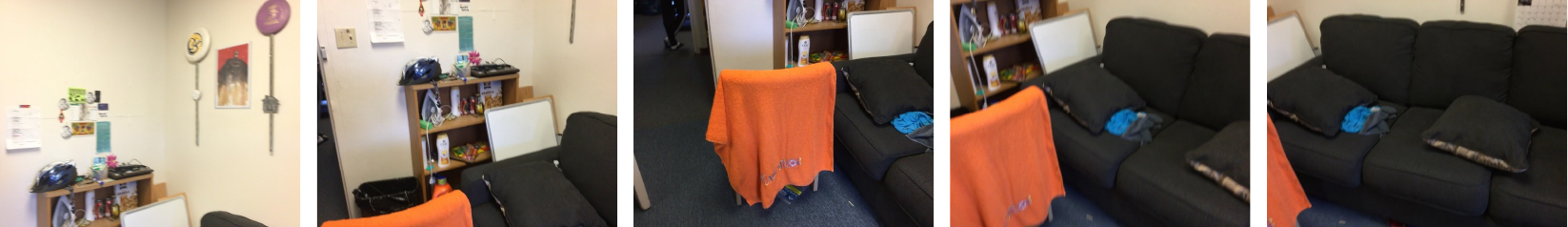}
    \caption{Failure case video due to VLM's limitation in generating spatial narrative.}
    \label{fig:fail1}
\end{figure}
\begin{figure}[h]
\centering
\includegraphics[width=0.98\linewidth]{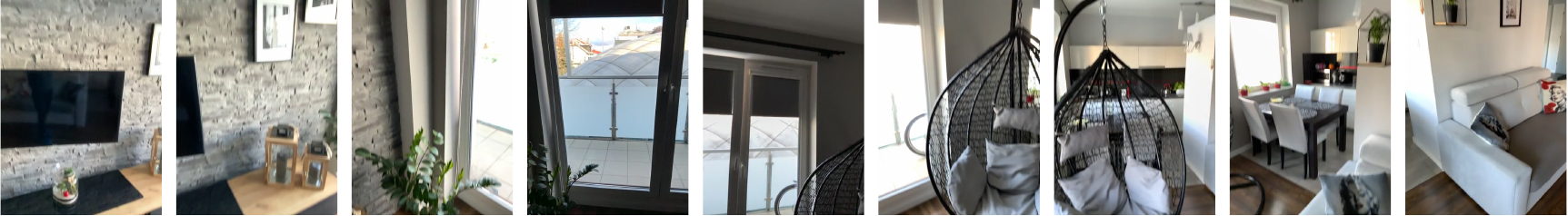}
    \caption{Failure case video with ambiguous and inaccurate ground truth.}
    \label{fig:fail2}
\end{figure}

\section{Failure Case Analysis}
\label{sec:failurecase}

We conduct extensive study on our failure cases, where we found most of the failure cases are caused by the VLM's inherent limitations and some of the ambiguous ground truth annotations. We discuss these two cases as follows.

\subsection{VLM Limitation}
In some failure cases, the VLM fails to generate accurate spatial narratives due to partial object visibility in the input video. Specifically, key objects required for answering the question may be occluded or appear only briefly, with the resulting spatial narrative lacks sufficient evidence, and further leads to incorrect prediction in SNS evaluation. These cases reflect inherent limitations of visual perception rather than deficiencies of the SNS framework. We illustrate an example in Figure~\ref{fig:fail1}, where several key objects~(chair, trash bin) are not mentioned in the VLM's spatial narrative, therefore the appearance order question is not answer correctly. The question and corresponding prediction and ground truth for failure case in Figure~\ref{fig:fail1} are as follows:

\begin{quote}
\textbf{Question.} What will be the first-time appearance order of the following categories in the video: table, backpack, window, chair?

\medskip
\textbf{SNS Prediction.} table, backpack, window, chair.

\medskip
\textbf{Ground Truth.} chair, table, backpack, window.
    
\end{quote}

\subsection{Ambiguous Ground Truth}
We also identify failure cases arising from ambiguity or incorrect in ground-truth annotations. In certain instances, the spatial question and corresponding answer is not precisely aligned with the video. Under SNS evaluation, both model-generated and human-written spatial narratives fail to recover the labeled answer.

The question and corresponding prediction and ground truth for failure case in Figure~\ref{fig:fail2} are as follows:

\begin{quote}
\textbf{Question.} Measuring from the closest point of each object, which of these objects (sofa, chair, stool, bathtub) is the closest to the tv?

\medskip
\textbf{SNS Prediction.} sofa.

\medskip
\textbf{Ground Truth.} chair.
    
\end{quote}

\section{Unsuccessful Attempts}

\subsection{Training with Automatic Generated Camera Motion Caption}
We initially explored leveraging ScanNet camera pose annotations to automatically construct camera motion caption as VLM supervision. Specifically, we converted raw camera poses into textual camera motion descriptions using pre-defined templates and performed supervised fine-tuning on these generated captions. However, we observed that camera motions derived from raw pose data are inherently noisy and exhibit severe distribution imbalance, with the majority of motions dominated by small horizontal panning (e.g., pan right). Furthermore, the exact camera motion can occasionally be hard to determined by the noisy, low-level camera pose. As a result, the generated textual data lacked diversity and precise video-language alignment. Training VLMs on this data led to rapid performance degradation and model collapse, as we found the VLM quickly perform shortcut learning by only generating amount-dominant camera motion. This suggests that naive conversion from raw pose signals to language descriptions is insufficient for stable VLM camera motion learning.

\subsection{Training with Raw Camera Pose Change Prediction}
We further attempted to directly supervise camera motion understanding by training the VLM to predict fine-grained camera pose changes, such as yaw, pitch, and roll deltas, from video input. In practice, this approach consistently resulted in training collapse. We find that current VLM architectures struggle to regress precise, continuous pose changes from visual input, especially when supervision requires fine-grained numerical accuracy. This indicates a mismatch between the representational capacity of language-centric models and the precision demanded by raw pose regression, highlighting the difficulty of integrating low-level geometric supervision into existing VLM frameworks. However, we believe introducing extra 3D reconstruction and camera estimation specialist models~\citep{wang2025vggt,wang2025continuous,wang2024dust3r} following recent works~\citep{zhao2025spacemind,fan2025vlm3r} can potentially improve the VLM learning process.

\section{Limitations and Future Work}
Our work focuses on learning and evaluating spatial understanding through narrative-based camera motion supervision, which emphasizes relational and qualitative spatial reasoning. While effective for diagnosing shortcut learning and improving camera motion understanding, this approach does not yet extend to precise metric or numerical spatial reasoning. Future work may explore hybrid representations that combine narrative-level spatial evidence with structured geometric signals, enabling models to reason across both qualitative and quantitative spatial domains.

In addition, SNS evaluates spatial narratives at the segment level and therefore does not directly assess a model’s ability to form a single, holistic representation of the global spatial layout across an entire video. This design choice is motivated by our assumption that reliable understanding of global spatial structure must be grounded in accurate interpretation of local camera motion. Without faithfully capturing local egomotion, reasoning over global spatial layout is unlikely to be reliable. Our ablation studies support this assumption: increasing the segment length consistently degrades spatial narrative quality, indicating that current VLMs struggle to accurately caption complex and long-horizon camera motions. This suggests that limitations in global spatial reasoning primarily stem from insufficient local motion understanding, and highlights an open challenge in enabling VLMs to scale from local egomotion comprehension to coherent global spatial modeling.


\section{Reproducibility Statement}
We have made extensive efforts to ensure the reproducibility of our work. All key implementation details, including model architecture, training procedures, and hyperparameter settings, are described in Table~\ref{table:training_config}. The evaluation benchmarks, data splits, and evaluation settings used in our experiments are detailed in the main paper and Section.~\ref{sec:implementation}. The prompt templates used for spatial narrative generation are provided in Section.~\ref{sec:template}. To facilitate full reproducibility, we will release the complete codebase for model training, evaluation, and deployment, along with model checkpoints, and the \datasetname~training dataset upon publication.

\end{document}